\documentclass[10pt, conference, compsocconf]{IEEEtran}
\ifCLASSINFOpdf
\else
\fi
\usepackage[pagebackref=true,breaklinks=true,letterpaper=true,colorlinks=true,bookmarks=true,bookmarksdepth=3]{hyperref}
\usepackage{url}  
\usepackage{times}                               
\usepackage{multirow}
\usepackage{latexsym}
\usepackage{epsfig}
\usepackage{multicol}
\usepackage{amsmath}
\usepackage{amssymb}
\usepackage{graphicx}
\usepackage{psfrag,color}
\usepackage{float}
\usepackage{tabularx}
\usepackage{amsmath}
\usepackage{listings}
\usepackage{multirow}

\usepackage{caption}
\usepackage{subcaption}
\usepackage{color}
\definecolor{dkgreen}{rgb}{0,0.6,0}
\definecolor{gray}{rgb}{0.5,0.5,0.5}
\definecolor{mauve}{rgb}{0.58,0,0.82}
\definecolor{pink}{rgb}{1,0.752941176,0.796078431}

\def\normalstretch{1.0}

\renewcommand{\baselinestretch}{\normalstretch}

\newcommand{\tab}{\hspace{0.35cm}}
\newcommand{\REM}[1]{}
\newcounter{step}

\usepackage{mathrsfs,amsmath} 
\usepackage{amssymb}  
\usepackage{algorithm}
\usepackage[noend]{algpseudocode}
\usepackage{epstopdf}

\begin{document}
%
\title{4-DoF Tracking for Robot Fine Manipulation Tasks}


\author{\IEEEauthorblockN{Mennatullah Siam*\thanks{ Authors contributed equally}, Abhineet Singh*, Camilo Perez and Martin Jagersand}
\IEEEauthorblockA{Faculty of Computing Science\\
University of Alberta, Canada\\
{\tt\small mennatul,asingh1,caperez,jag@ualberta.ca}} 
}


%


\maketitle

\begin{abstract}
This paper presents two visual trackers from the different paradigms of learning and registration based tracking and evaluates their application in image based visual servoing. They can track object motion with four degrees of freedom (DoF) which, as we will show here, is sufficient for many fine manipulation tasks.
One of these trackers is a newly developed learning based tracker that relies on learning discriminative correlation filters while the other is a refinement of a recent 8 DoF RANSAC based tracker adapted with a new appearance model for tracking 4 DoF motion. 

Both trackers are shown to provide superior performance to several state of the art trackers on an existing dataset for manipulation tasks. Further, a new dataset with challenging sequences for fine manipulation tasks captured from robot mounted eye-in-hand (EIH) cameras is also presented. These sequences have a variety of challenges
encountered during real tasks including jittery camera movement, motion blur, drastic scale changes and partial occlusions. Quantitative and qualitative results on these sequences are used to show that these two trackers are robust to failures while providing high precision that makes them suitable for such fine manipulation tasks. 
\end{abstract}

\renewcommand\footnotemark{}
\renewcommand\footnoterule{}

%
\IEEEpeerreviewmaketitle

\section{INTRODUCTION}
2D Object tracking is a core component in visual servoing~\cite{hutchinson1996tutorial} where visual feedback is used to guide the robot to perform certain tasks. One category of these tasks involves manipulation of objects \cite{gridseth2016vita} ranging from simple pick and place to more advanced ones. Fine manipulation \cite{quintero2014small} in particular, where small objects are handled, can be quiet challenging. Though a lot of research has been done using depth cameras like Kinect for manipulation tasks in general, yet these sensors are not suitable for fine manipulation. This is due to limitations in their range, resolution and accuracy. Most current depth cameras have an operation range of 0.8 to 5 m which is not enough for EIH configurations \cite{quintero2014small}. Some have used a Kinect sensor for initial positioning, followed by manual tele-operation to grasp objects \cite{jiang2013integrated}. However, a more versatile solution is to use image based visual servoing (IBVS) by itself for such scenarios.
\begin{centering}
\begin{figure}[t!]
 \includegraphics[width=\linewidth]{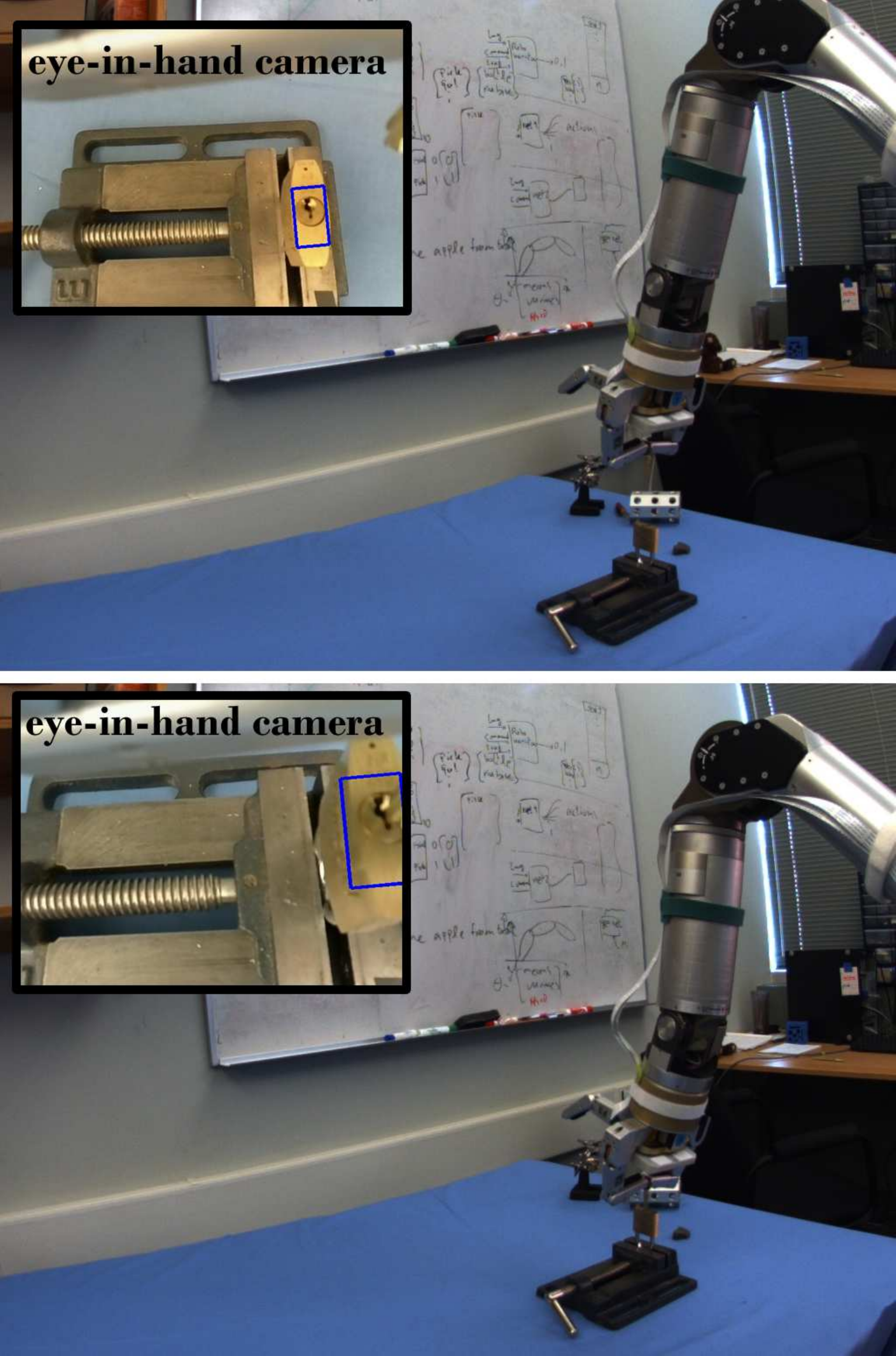}

\caption{Fine manipulation task of inserting key into lock. Images from eye-to-hand and eye-in-hand configuration}
\label{fig:setup}
\end{figure}
\end{centering}

\par In this paper, we provide a solution to visual tracking for performing fine manipulation tasks using IBVS with low cost cameras. The presented trackers are shown to work well with EIH configuration within very small ranges and with better accuracy than several state of the art trackers. They provide 4 DoF motion information, sufficient for high precision tasks to work, while being robust to occlusions, illumination changes and motion blur. We have integrated them as part of a tracking library called Modular Tracking Framework (MTF) \cite{singh2016modular} along with all trackers tested here so the results are easily reproducible. MTF also provides a ROS interface for effortless integration in robotics applications.\\

To summarize, following are the main contributions of this work:
\begin{itemize}
\item A new 4 DoF correlation based tracker Rotation and Scale Space Tracker (RSST) is introduced.
\item A state of the art RANSAC based tracker (RKLT) \cite{zhang2015rklt} is adapted to be more robust to appearance changes by incorporating an illumination invariant similarity metric and limiting it to 4 DoF motion estimation.
\item  A new benchmark \cite{tmt} is made publicly available with 24 video sequences captured from both a fixed camera in eye-to-hand (ETH) configuration and a robot end effector mounted EIH stereo camera that are annotated (Fig. \ref{fig:setup}). We call it Tracking for Fine Manipulation Tasks (\textbf{TFMT}) dataset.
\item Detailed experimental analysis for these trackers is presented in the context of fine manipulation tasks. Several new insights are obtained and discussed regarding the differences between tracking for manipulation tasks and general 2D object tracking. Experimental results that motivate our choice of 4 DoF are provided showing its advantages over both lower and higher DoF tracking.
\end{itemize}
\par Rest of this paper is organized as follows: section \ref{sec:rel} provides an overview of the related work in 2D visual tracking followed by details about the two trackers presented here in sections \ref{sec:rsst} and \ref{sec:rklt}. Section \ref{sec:exps} presents the experimental analysis, first on general manipulation tasks, then on specific fine manipulation tasks. This is followed by tests on general object tracking scenarios. Finally section \ref{sec:conc} presents the conclusions.

\section{RELATED WORK}
\label{sec:rel}
The general 2D object tracking problem has been extensively researched in past decades. Trackers can be categorized as discriminative trackers and generative. In the former category \cite{hare2011struck}\cite{kalal2012tracking}\cite{zhang2012real}, the tracking task is posed as a binary classification problem. A discriminative classifier is then learned online from patches containing the object and the background. These learning based approaches are able to cope to some extent with illumination variations, partial occlusions, and viewpoint changes. 
\par Generative trackers, on the other hand, learn a model to represent the object and then use it to search the current frame for the object. They can learn the model online \cite{ross2008incremental} or can have a static model as in most registration based trackers \cite{baker2004lucas}\cite{benhimane2004real}. These latter, in a sense, represent a different paradigm of object tracking where precise pose of the object is needed as opposed to a rough bounding box. Their goal is to estimate the optimal warp parameters between the current patch and the reference one. Since several of these use gradient based methods for computing the warp parameters,  they are also computationally efficient. However, they often tend to fail under occlusions or other appearance changes, working, as they do, under the assumption that changes in the appearance are solely due to the warping. A relatively recent tracker in \cite{zhang2015rklt} combined a set of simple 2 DoF Lucas Kanade (LK) feature trackers with RANSAC to estimate 8 DoF motion. RANSAC rejected lost trackers as outliers thus increasing its robustness. It, however, used the sum of squared differences (SSD) of image intensities as the similarity measure which made it vulnerable to failure in the presence of illumination changes and partial occlusions. Also, its use of 8 DoF motion model made it more prone to getting stuck in local optima.
\par In recent research, trackers using discriminative correlation based filters \cite{henriques2015high}\cite{danelljan2014accurate}\cite{bolme2010visual} have shown great success. However, these trackers tend to provide DoF motion information where only translation is computed though some also estimate isotropic scaling \cite{kalal2012tracking}\cite{danelljan2014accurate}. However, for manipulation tasks, knowledge of the \textit{orientation} of the object is necessary to be able to guide the robot motion precisely. The CMT tracker \cite{nebehay2015clustering} incorporates rotation but it relies heavily on the detected key points and their descriptors and thus faces difficulties with less textured objects that are common in industrial scenarios. Recently, a deep regression network \cite{held2016learning} was used to track by matching the query template within a candidate region. Although it was shown to run at 100 frames per second with offline training on a large dataset, it does not generalize well enough to variations in the sequences as will be demonstrated for manipulation scenarios.
\par In terms of tracking evaluation, the general object tracking category has two recent benchmarks - VOT \cite{vot16} and OTB \cite{wu2013online} - that overlap in some of the sequences. However, these benchmarks are not suitable for robotics applications as they predominantly feature surveillance type videos and their ground truth is also not very precise. A recent tracking benchmark for manipulation tasks \cite{tmt} provided a public dataset for robotics scenarios. It had several challenges including partial occlusions, out-of-plane rotation and illumination changes. Nonetheless, it lacked sequences that encompass complete tasks and also did not have any that were captured from EIH configuration that can cause great variability in scale. Finally, the motion - both by human and robotic arm - in all of its sequences was executed too smoothly to accurately represent realistic tasks where the motion is often jerky.

\section{Rotation and Scale Space Tracker (RSST)}
\label{sec:rsst}
The proposed approach is closely related to \cite{henriques2015high}\cite{danelljan2014accurate}\cite{bolme2010visual}. These trackers are based on learning a discriminative correlation filter to localize the object of interest. Some of the above mentioned trackers support only 2D translation  \cite{henriques2015high}\cite{bolme2010visual} while others were extended to include isotropic scaling \cite{danelljan2014accurate}. In this section, these correlation based trackers are extended further to include rotation. Similar to \cite{danelljan2014accurate}, HOG features of the search region patch are used and denoted as $x$. Assuming that this feature map is of dimension $d$, feature map $l\in{1,...,d}$ is denoted as $x^l$. A correlation filter $h^l$ is then learned for each feature dimension by optimizing the following objective function:
\begin{equation}
C= \lVert \sum_{l=1}^d h^l*x^l- f \rVert^2 + \lambda  \sum_{l=1}^d \lVert h^l \rVert^2
\end{equation}
where $f$ is the desired correlation output and $\lambda$ is a factor to control the regularization term. The desired correlation output is a Gaussian centered at the optimum translation, scale or rotation. Solving the above equation in the frequency domain yields:
\begin{equation}
H^l= \frac{F\bar X^l}{ \sum_{k=1}^d X^k \bar X^k + \lambda}
\end{equation}
where $H$, $F$, $X$ denote the discrete Fourier transforms of their corresponding signals $h$, $f$ ,$x$, and $\bar X$ is the complex conjugate. The power of correlation based tracking is its usage of the simple convolution theorem to formulate the problem in the Fourier domain. This approach makes it possible to learn a linear classifier for different shifts of the original patch without rigorously going through all of them as in other tracking by detection approaches.

\par In order to adapt to appearance changes of the object, the correlation filter is updated according to the following equations:
\begin{equation}
N^l_t= (1- \eta) N^l_{t-1} + \eta F_t\bar X^l_t 
\label{eq1}
\end{equation}
\begin{equation}
D_t= (1- \eta) D_{t-1}+ \eta \sum_{k=1}^d X_t^k \bar X_t^k
\label{eq2}
\end{equation}
where  $N^l_t$ and $D_t$ respectively denote the numerator and denominator of the correlation filter for feature dimension $l$ at time instant $t$ while $\eta$ is the learning rate. This mechanism ensures that the tracker does not drift with each update as it relies on previous history as well. Finally, the optimum parameter - whether it is for translation, scale or rotation - is computed from the peak response of the correlation between new feature maps and the correlation filter.
\begin{equation}
y= \mathscr{F}^{-1}\{ \frac{\sum_{l=1}^d N^l Z^l}{D + \lambda} \}
\label{eq3}
\end{equation}
where $y$ is the new parameter and $Z$ denotes the discrete Fourier transform of the feature maps of the new candidate region.
\par An ideal rotation and scale space tracker will search through the joint space of translation, rotation and scale. However, for the sake of computational efficiency, separate correlation filters for translation, scale and rotation are learned instead. This choice is based on the experiments in \cite{danelljan2014accurate} that compared the joint computation of translation and scale against separate ones and showed that the latter provided a significant advantage in computational efficiency without degradation in accuracy. For computing the orientation, rotation samples are extracted within a range of potential rotations $ [-20,20] $ with $ 2 $ degrees increment. A patch is extracted for each candidate rotation and HOG features are used to encode it. This constructs the rotation feature matrix that is used within the correlation equations. This is used afterwards to update the components of a 1D correlation filter. 
The steps used for updating the tracker at each time step $t$ are shown in Algorithm \ref{alg:rsst}.
\begin{algorithm}
\caption{Rotation and Scale Space Tracker}\label{euclid}
\label{alg:rsst}
\begin{algorithmic}[1]
\Statex \textbf{Input:}
\Statex \tab Image $I_t$
\Statex \tab Target position $p_{t-1}$, scale $s_{t-1}$ and rotation $r_{t-1}$.
\Statex \tab Translation model $N^{trans}_{t-1}$, $D^{trans}_{t-1}$
\Statex \tab Scale model $N^{scale}_{t-1}$, $D^{scale}_{t-1}$
\Statex \tab Rotation model $N^{rot}_{t-1}$, $D^{rot}_{t-1}$
\Statex
\Statex \textbf{Output:}
\Statex \tab New target position $p_{t}$, scale $s_{t}$ and rotation $r_{t}$.
\Statex \tab Models $N^{trans}_{t}$, $D^{trans}_{t}$,$N^{scale}_{t}$, $D^{scale}_{t}$,$N^{rot}_{t}$, $D^{rot}_{t}$
\Statex
\Statex \textbf{Translation}
\State Extract features for translation sample $z_{trans}$ at $p_{t-1}$, $s_{t-1}$, $r_{t-1}$.
\State Compute translation response $y_{trans}$ using $z_{trans}$, $N^{trans}_{t-1}$, $D^{trans}_{t-1}$ as in equation \ref{eq3}.
\State Set $p_t$ to maximum location of 2D response $y_{trans}$.
\Statex
\Statex \textbf{Scale}
\State Extract features for scale sample $z_{scale}$ at $p_{t}$, $s_{t-1}$, $r_{t-1}$.
\State Compute scale response $y_{scale}$ using $z_{scale}$, $N^{scale}_{t-1}$, $D^{scale}_{t-1}$ similar to translation.
\State Set $s_t$ to maximum location of 1D response $y_{scale}$.
\Statex
\Statex \textbf{Rotation}
\State Extract multiple patches at different sampled rotations in the range [-20,20] around the previous accumulated rotation.
\State Extract HOG features for rotation sample $z_{rot}$ at $p_{t}$, $s_{t}$, $r_{t-1}$. 
\State Compute rotation response $y_{rot}$ using $z_{rot}$, $N^{rot}_{t-1}$, $D^{rot}_{t-1}$ similar to translation . 
\State Set $r_t$ to maximum location of 1D response $y_{rot}$.
\Statex
\Statex \textbf{Update}
\State Extract samples $x_{trans}$, $x_{scale}$, $x_{rot}$ at the new parameters.
\State Compute $N^{trans}_t$, $D^{trans}_{t}$,$N^{scale}_{t}$, $D^{scale}_{t}$,$N^{rot}_{t}$, $D^{rot}_{t}$ with equations \ref{eq1} and \ref{eq2}.
\end{algorithmic}
\end{algorithm}

\section{Ransac Based Tracker (RKLT)}
\label{sec:rklt}
\par This is a state of the art registration based tracker \cite{zhang2015rklt} that is also used in the experiments to demonstrate the benefits of 4 DoF tracking and contrast the two trackers from different paradigms that provide solutions to the same problems in the fine manipulation context.
RKLT is a two layer tracker. In the first layer, evenly sampled points are tracked between consecutive images using the pyramidal KLT tracker\cite{bouguet2001pyramidal} and the corresponding point pairs are used as input to a RANSAC based method that estimates that similarity transform that can best explain the warping between them.
The points that could not be tracked by the KLT tracker, due to partial occlusions or other appearance changes, are rejected as outliers and not used for the RANSAC estimation. 

\par
The output of the first layer is used as input to an inverse compositional (IC) Lucas Kanade tracker \cite{baker2001equivalence} that refines it further by aligning it with the original template.
Only inliers are considered in the IC algorithm to achieve better convergence.
Finally, in order to provide robustness to illumination changes, normalized cross correlation (NCC) \cite{Scandaroli2012_ncc_tracking} is used as the similarity metric in the second layer rather than the conventional SSD \cite{baker2001equivalence}.

\section{Experimental Analysis}
\label{sec:exps}
This section details the experimental setup and data collection procedure.
It also presents quantitative analysis of the two trackers compared again eight state of the art trackers from literature. The results are first presented for general manipulation tasks using the recent tracking benchmark for manipulation tasks (TMT) \cite{trackerManipulation}. This is followed by evaluation on the specific fine manipulation aspect on the TFMT dataset this is presented in this work. Finally an evaluation for the general 2D object tracking using VOT\cite{vot16} benchmark is also provided to highlight the differences in the two domains of tracking.

\subsection{Experimental Setup}
TFMT includes sequences with three fine manipulation tasks performed through tele-manipulation:
1) Opening a lock with a key (denoted as Key Task in our experiments).
2) Inserting a thread through a fishing lure (denoted as Fish Lure).
3) Inserting a rivet in an industrial part (denoted as Hexagon Task).
Our data collection was performed through tele-manipulation using a 4-DoF arm with a hand gimbal as master and a 7-DoF WAM arm  with a barrett hand as slave. This gives a different type of motion compared to the ones in TMT sequences that were executed by smooth human and robot motions. 

\begin{figure}
      \centering
      \includegraphics[width=0.46\textwidth]{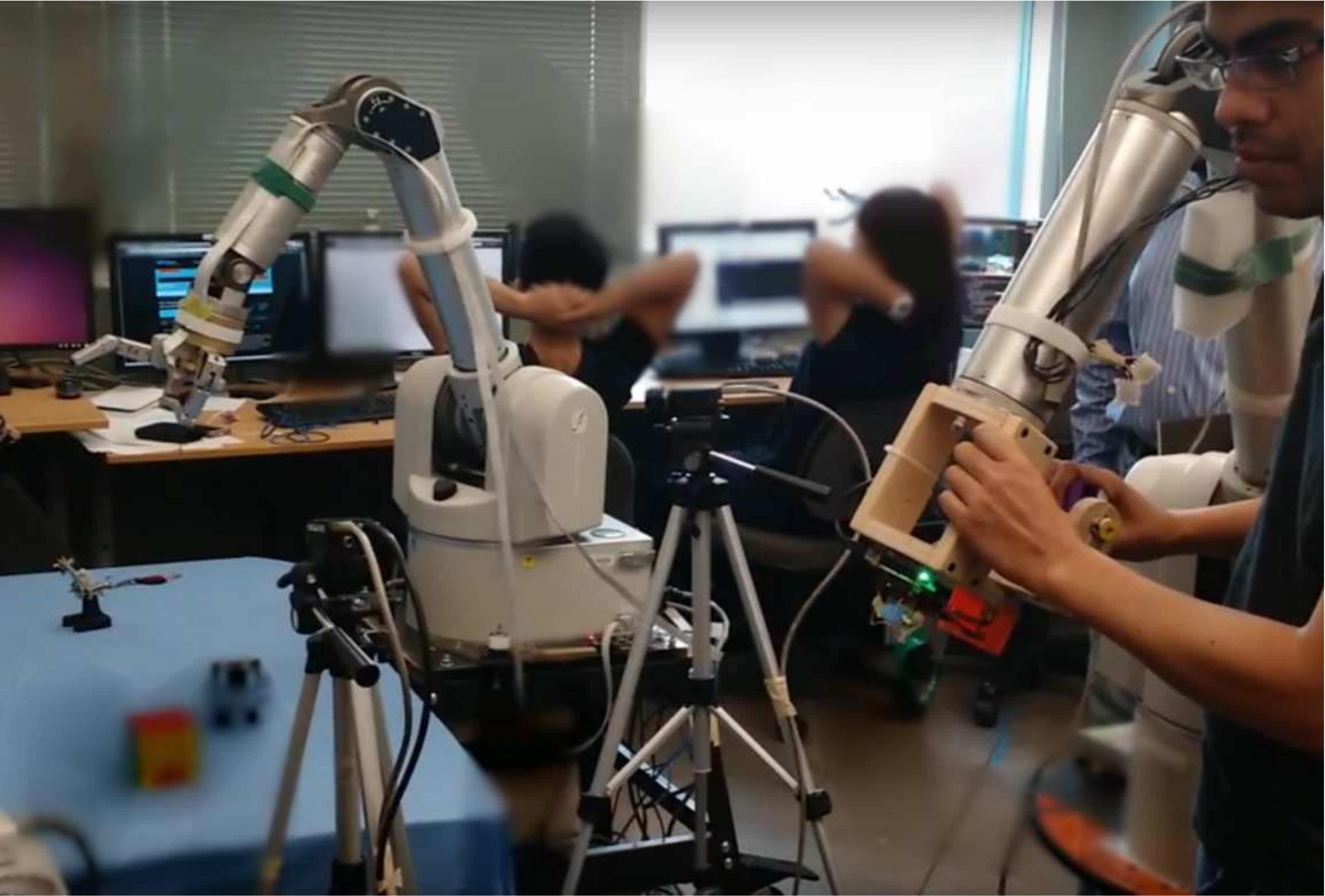}
      \caption{ Tele-manipulation setup used for experimental data collection. Left: 7-DoF WAM arm with barrett hand and two eye-inhand-cameras. Right: 4-DoF WAM arm with hand gimbal. 
}
      \label{fig:teleop}
   \end{figure}

The experimental setup is shown in Figure \ref{fig:teleop}. We used two raspberry pi cameras located in the barrett hand for the eye-in-hand configuration and two point gray grass hopper cameras with 3mm lenses for the eye to hand configuration. The resolution of the images recorded from eye-in-hand configuration is 640x480. Since two cameras were used, TFMT has 12 sequences in all with a total of 1841 frames. In order to better utilize the frames that follow a tracker's first failure in any sequence, subsequences were used during evaluation where the tracker was initialized at 10 different frames. This increases the effective number of frames to 10,360. The sequences are made publicly available \cite{tmt}.   

\par The challenges posed in these sequences include partial occlusions, and objects partially going out of the field of view. Motion blur was a significant challenge as well, especially in the fast version of these sequences. Some of the objects were texture less like the industrial aluminum part. It is worth noting that one of the main differences to the TMT dataset \cite{trackerManipulation} is that these sequences are captured with eye-in-hand configuration. 
The images in TMT had to be resized to half the size for RSST in order to obtain real-time performance. However, in fine manipulation tasks, since the eye-in-hand cameras have low resolutions, the original size is maintained.  

\par We compare against 8 state of the art trackers: TLD \cite{kalal2012tracking}, DSST \cite{danelljan2014accurate}, CMT \cite{nebehay2015clustering}, KCF \cite{henriques2015high}, Struck \cite{hare2011struck}, Fragtrack \cite{adam2006robust} and Goturn \cite{held2016learning} and RCT \cite{zhang2012real}. The error metric used for evaluating the tracking performance in the manipulation tasks experiments is the alignment error $E_{al}$:
\begin{equation}
\label{eq:malerror}
E_{al}= \sqrt{ \frac{\sum_{i=1}^4 (X_{T_i}- X_{GT_i})^2 }{4} }
\end{equation}
where $X_T$ denotes the tracking output corners, and $X_{GT}$ is the corresponding ground truth. The reason for using this metric is that manipulation tasks rely heavily on the accuracy of the measurements.
For VOT evaluation, however, Jaccard error  $E_{jac}$ (eq \ref{eq:jaccerror}) is used following the exact procedure in \cite{vot16} since it is sufficient for the scenarios in that dataset:
\begin{equation}
\label{eq:jaccerror}
E_{jac}= 1- \frac{ A_T \cap A_{GT} }{ A_T \cup A_{GT} }
\end{equation}
where $A_T$ is the tracking output bounding box, and $A_{GT}$ is the ground truth. 

\subsection{Tracking for Manipulation Tasks}
\begin{figure*}[!htbp]
\begin{center}
\includegraphics[width=\textwidth]{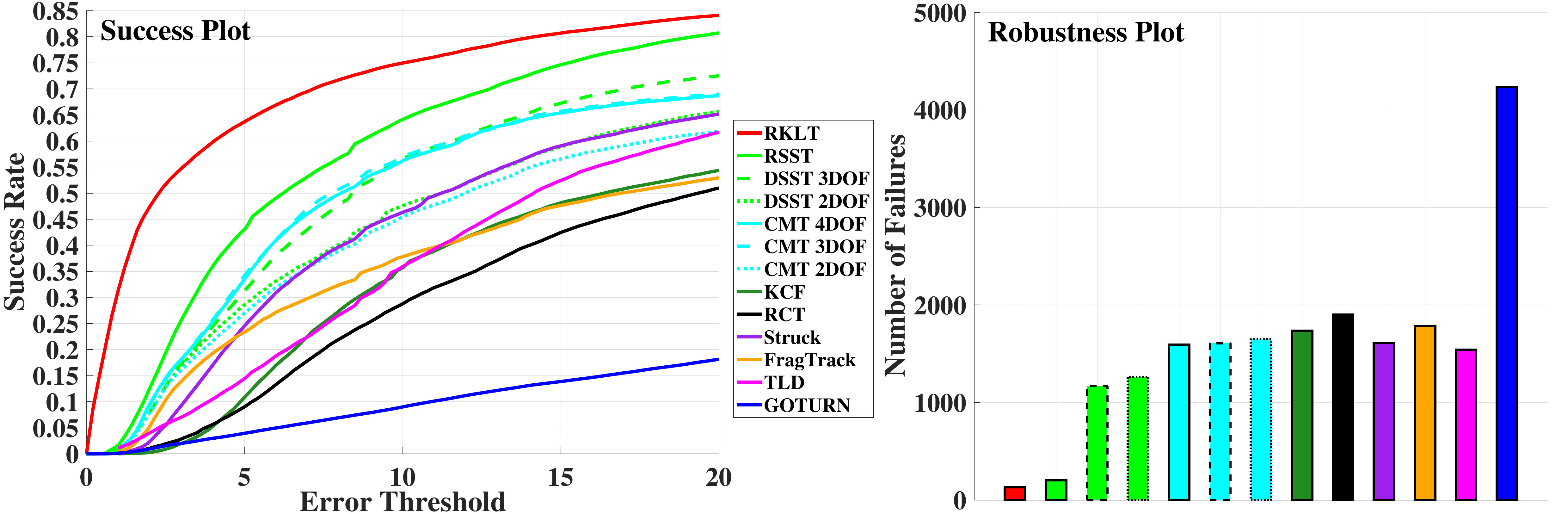}
\caption{Comparing different trackers on TMT using alignment error(MCD error), where RKLT and RSST outperform the state of the art.}
\label{fig:tmt_mcd}
\end{center}
\end{figure*}

\begin{figure*}[!htbp]
\begin{subfigure}[b]{0.25\textwidth}
                \includegraphics[width=\linewidth]{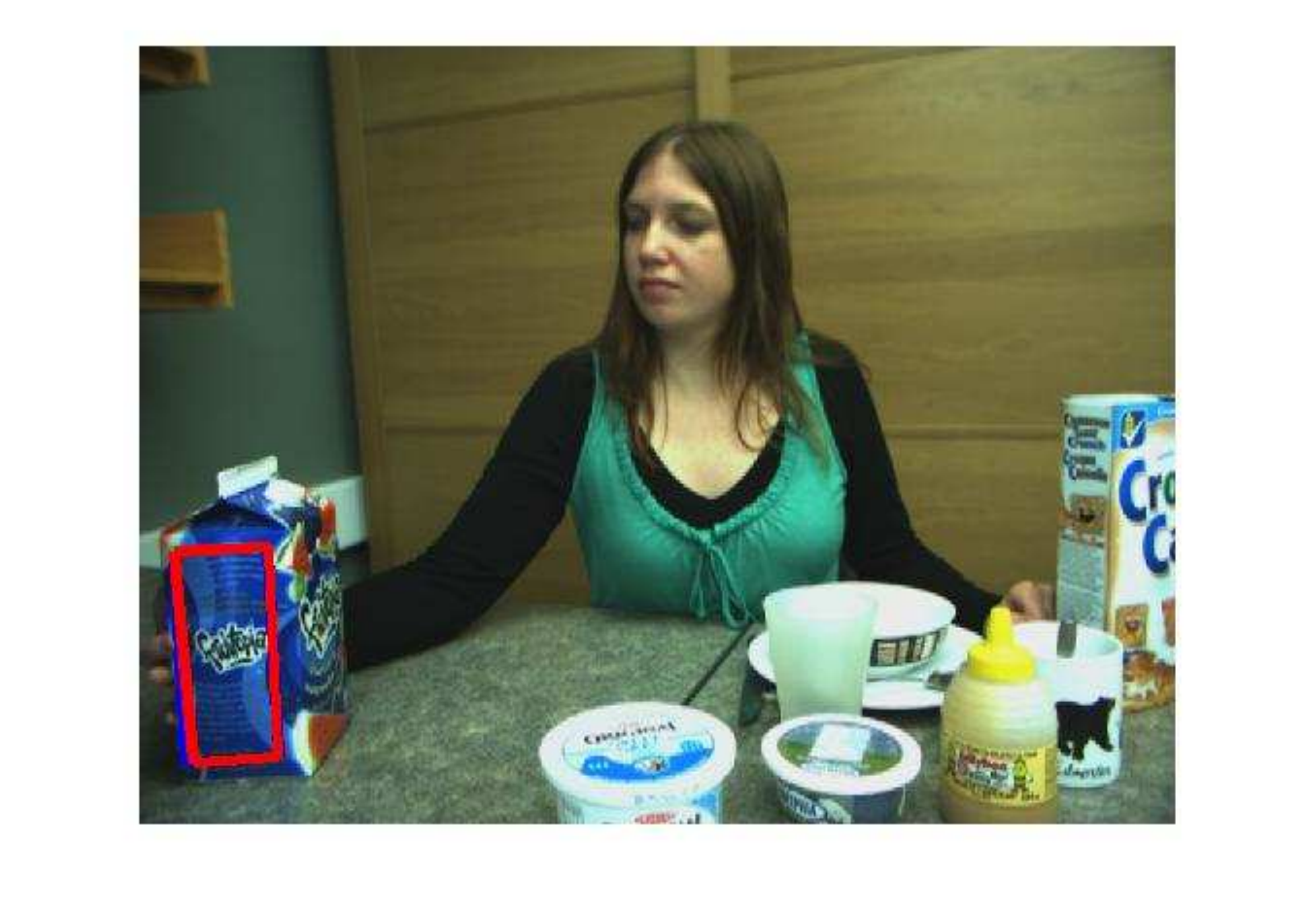}
        \end{subfigure}%
        \hspace{-1\baselineskip}
        \begin{subfigure}[b]{0.25\textwidth}
                \includegraphics[width=\linewidth]{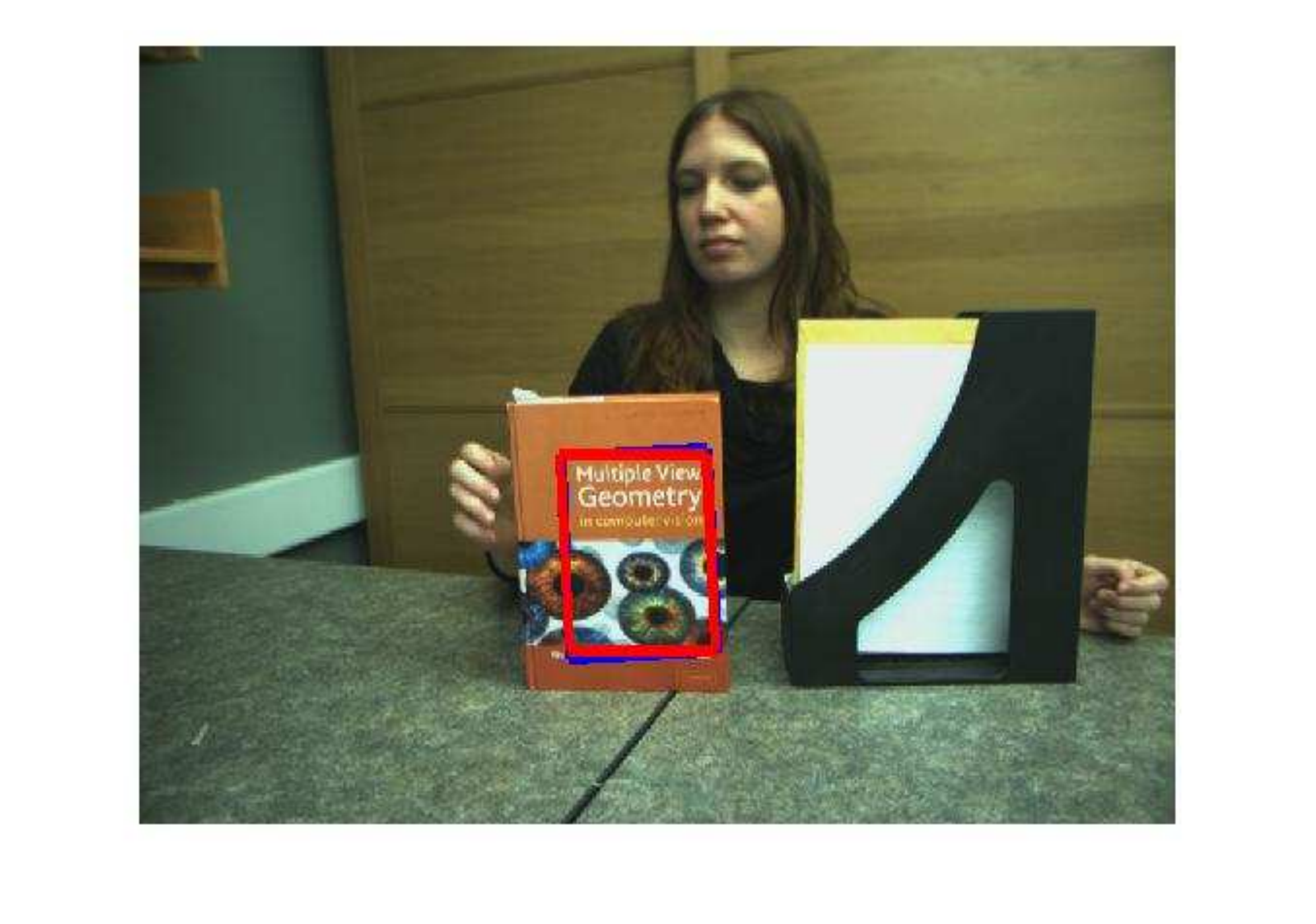}
        \end{subfigure}%
        \hspace{-1\baselineskip}
        \begin{subfigure}[b]{0.25\textwidth}
                \includegraphics[width=\linewidth]{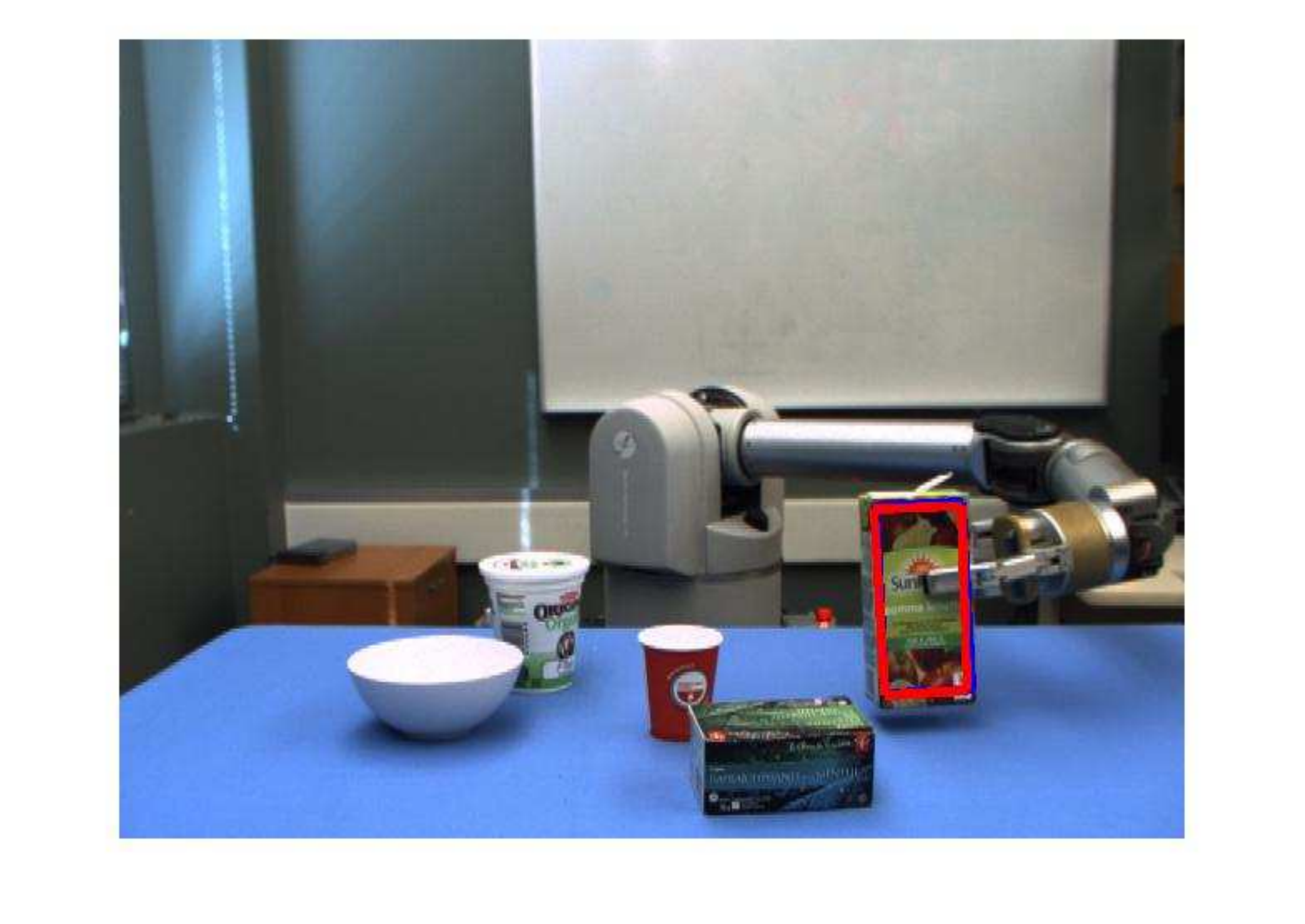}
        \end{subfigure}%
        \hspace{-1\baselineskip}
        \begin{subfigure}[b]{0.25\textwidth}
                \includegraphics[width=\linewidth]{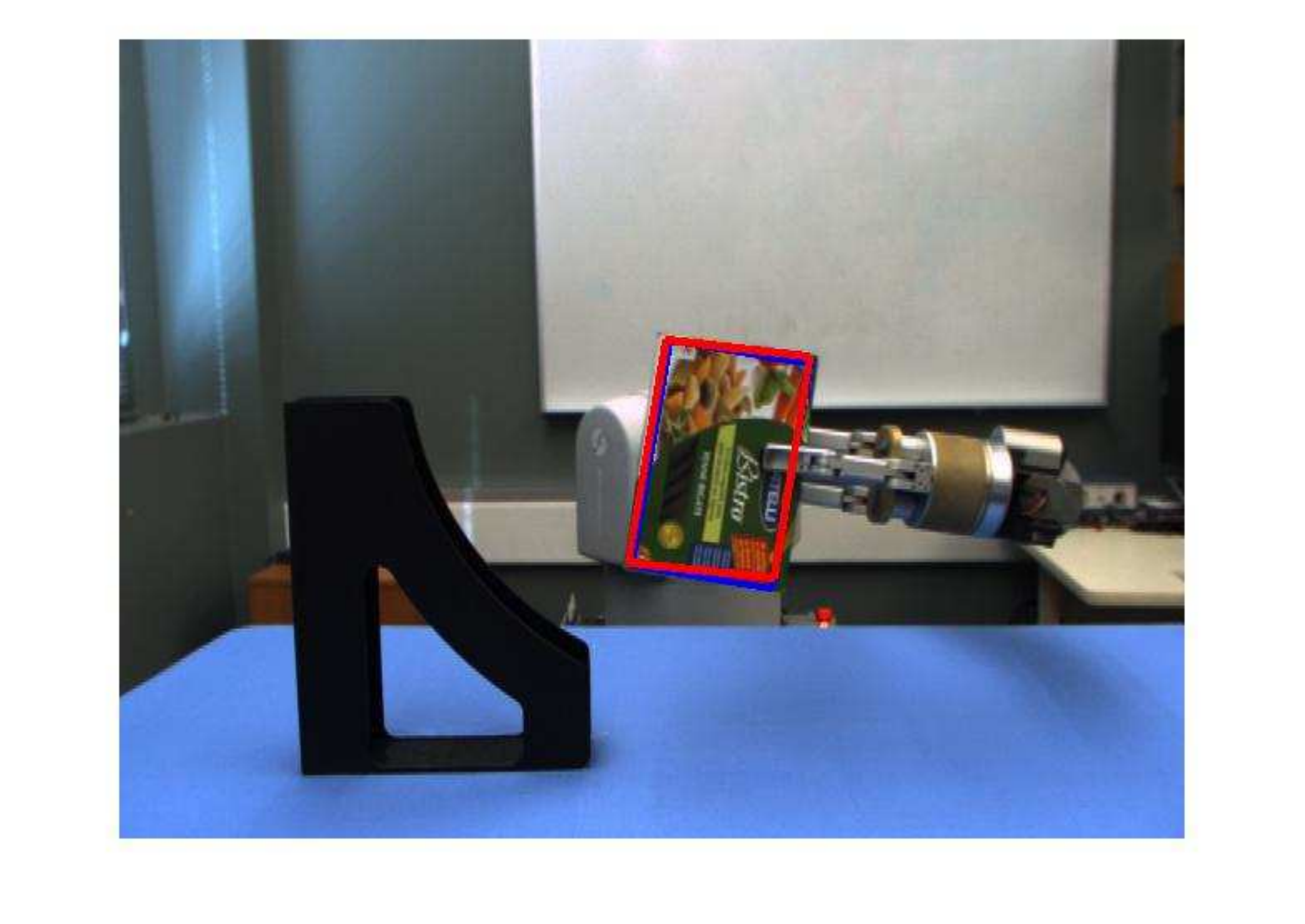}
        \end{subfigure}
        
       
        \vspace{-1\baselineskip}
        \begin{subfigure}[b]{0.25\textwidth}
          \includegraphics[width=\linewidth]{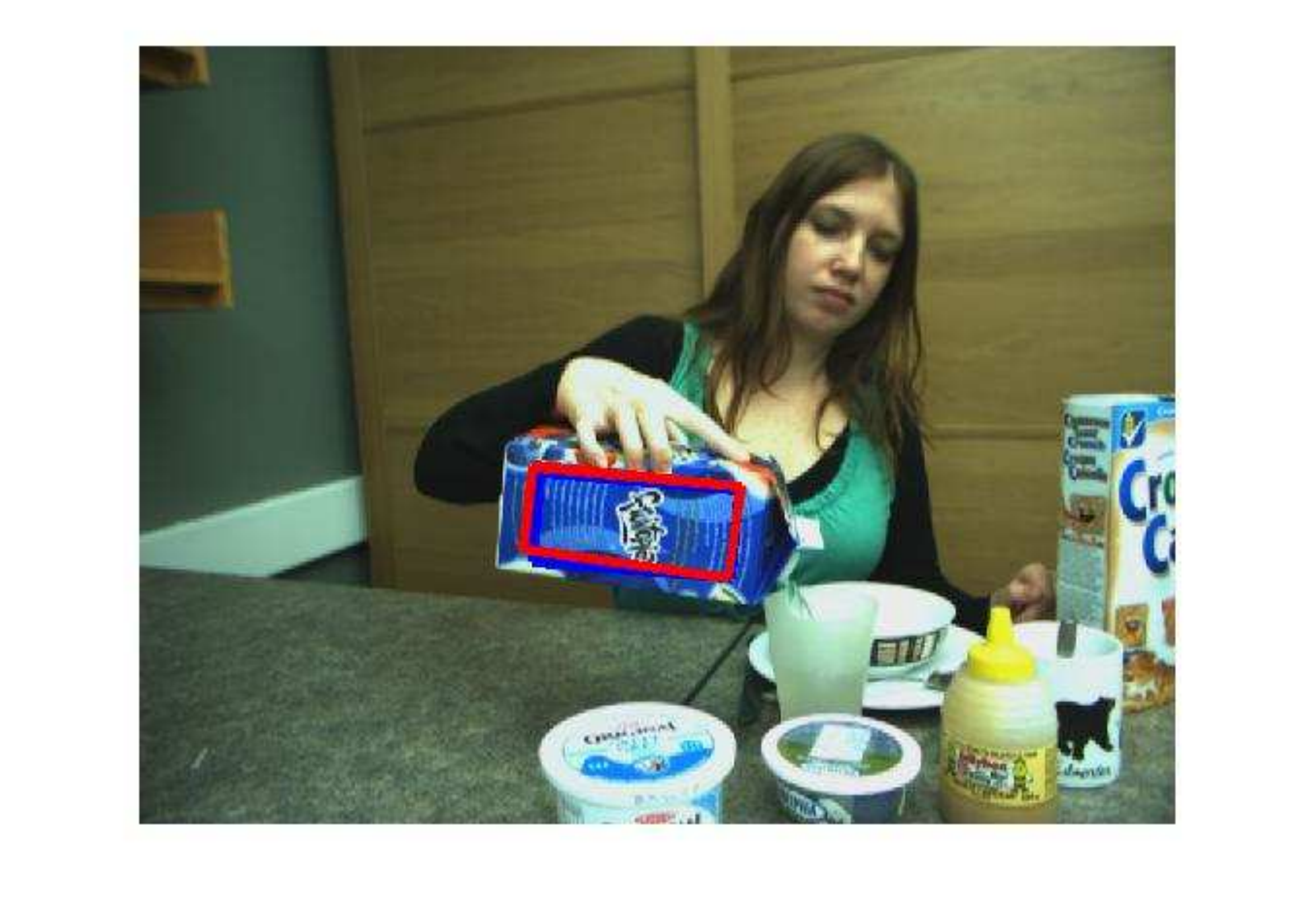}
         \subcaption{}
        \end{subfigure}%
        \hspace{-1\baselineskip}
        \begin{subfigure}[b]{0.25\textwidth}
                \includegraphics[width=\linewidth]{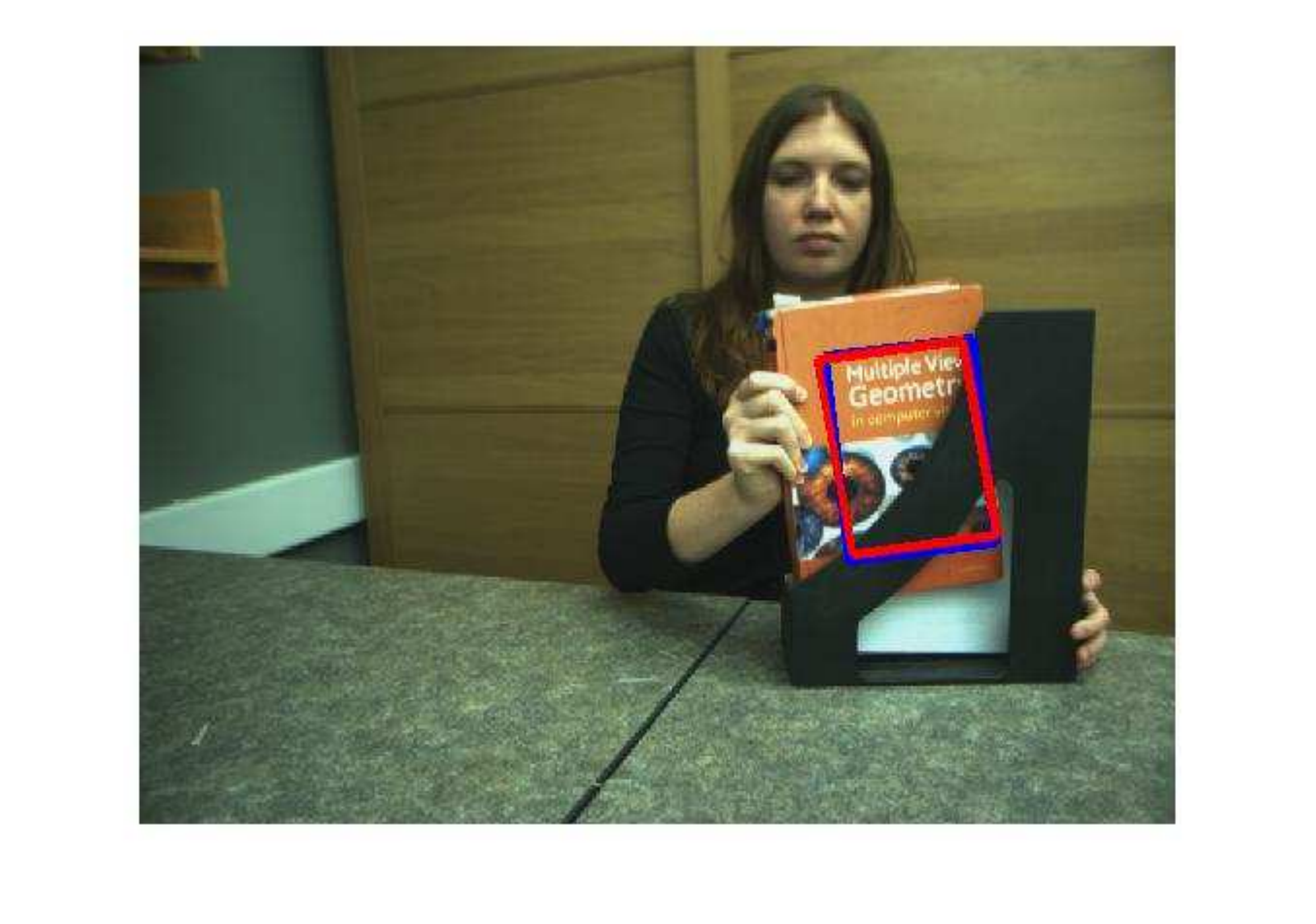}
         \subcaption{}
        \end{subfigure}%
        \hspace{-1\baselineskip}
        \begin{subfigure}[b]{0.25\textwidth}
                \includegraphics[width=\linewidth]{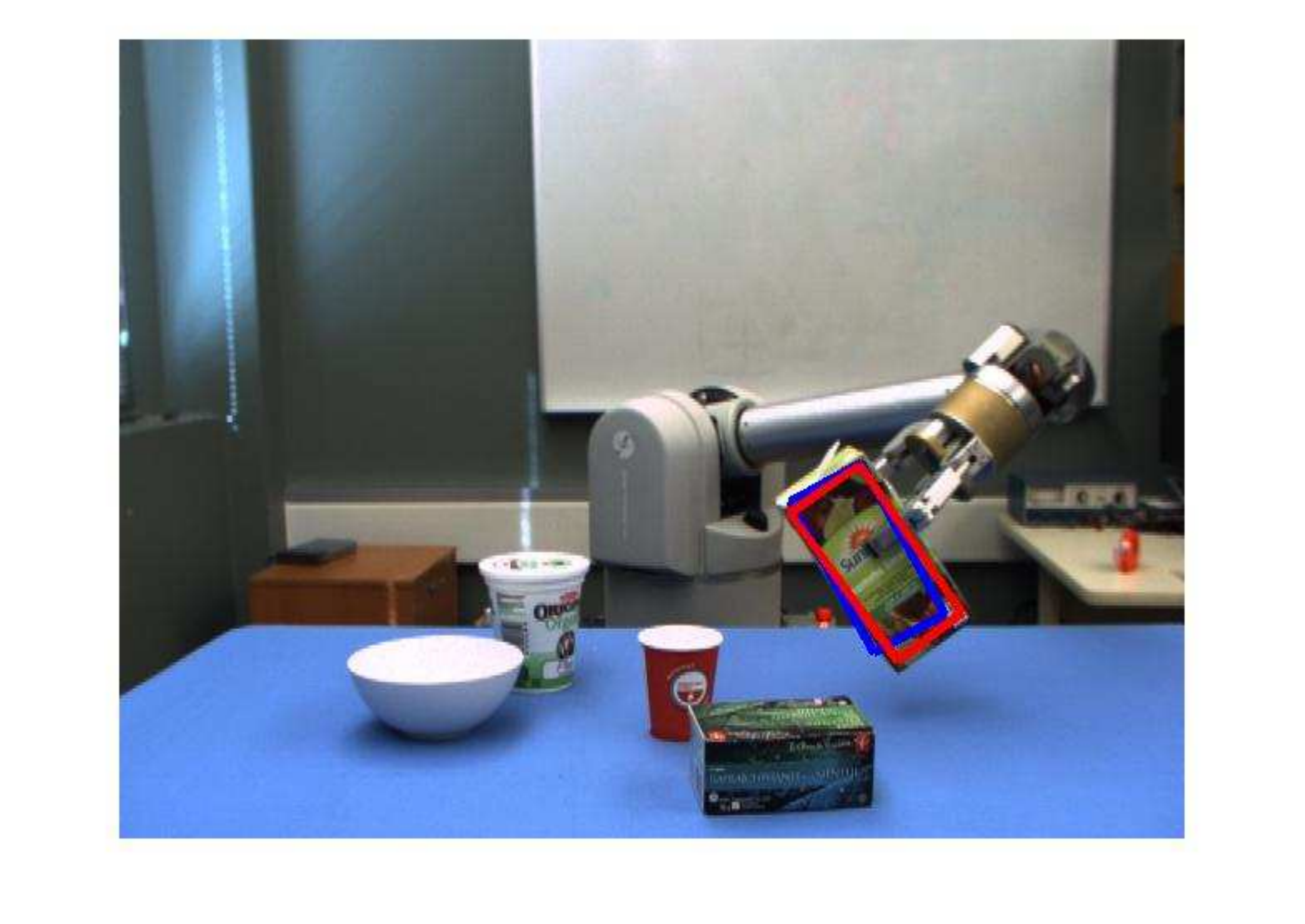}
        \subcaption{}
        \end{subfigure}%
        \hspace{-1\baselineskip}
        \begin{subfigure}[b]{0.25\textwidth}
                \includegraphics[width=\linewidth]{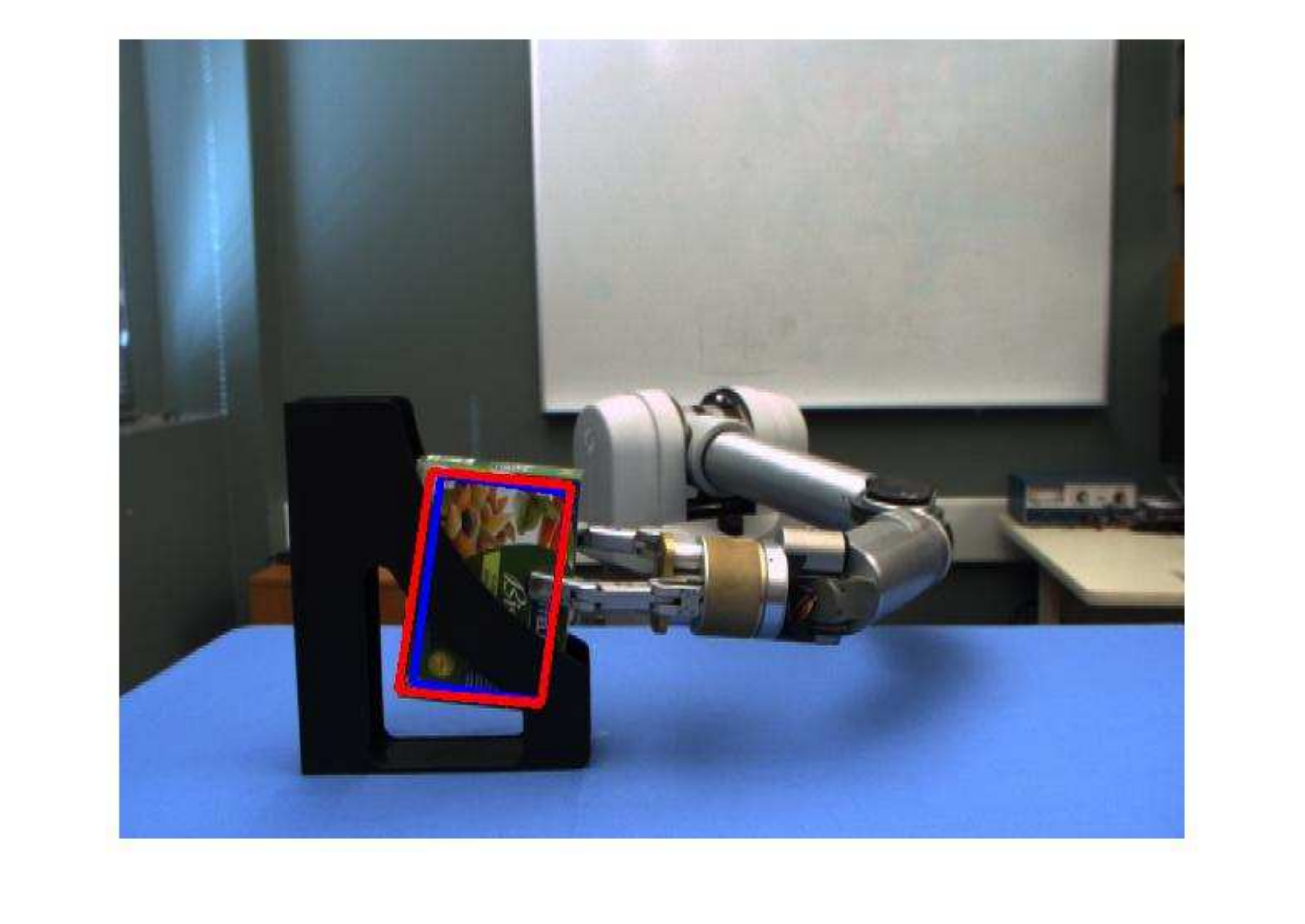}
        \subcaption{}
        \end{subfigure}
         
\caption{Tracking Results for RSST in blue and RKLT in red on TMT. a: Juice Sequence. b: Book III Sequence. c: Robot Juice Sequence. d: Robot BookIII Sequence}
\label{fig:tracktmt}
\end{figure*}

This section presents the quantitative and qualitative evaluation on the manipulation tasks benchmark in \cite{trackerManipulation}. Success and robustness plots are shown in Figure \ref{fig:tmt_mcd}. It can be seen that RKLT and RSST both significantly outperform all other trackers with respect to both robustness and accuracy. It can be seen too that RKLT which is a registration based tracker outperforms RSST in terms of accuracy since gradient based methods that are used in registration based trackers tend to provide higher precision. 

Qualitative results for these two trackers in the general manipulation tasks are shown in Figure \ref{fig:tracktmt}. The first and third columns show scenarios where estimating rotation of the object tracked is necessary to track it accurately. RSST and RKLZT are both capable of tracking the rotated object with RKLT providing more precise results. The second and last columns show scenarios of partial occlusions. Again RSST and RKLT are both able to track the occluded object robustly.

\subsection{Tracking for Fine Manipulation (TFMT)}
\begin{figure*}[!htbp]
\begin{center}
\includegraphics[width=\textwidth]{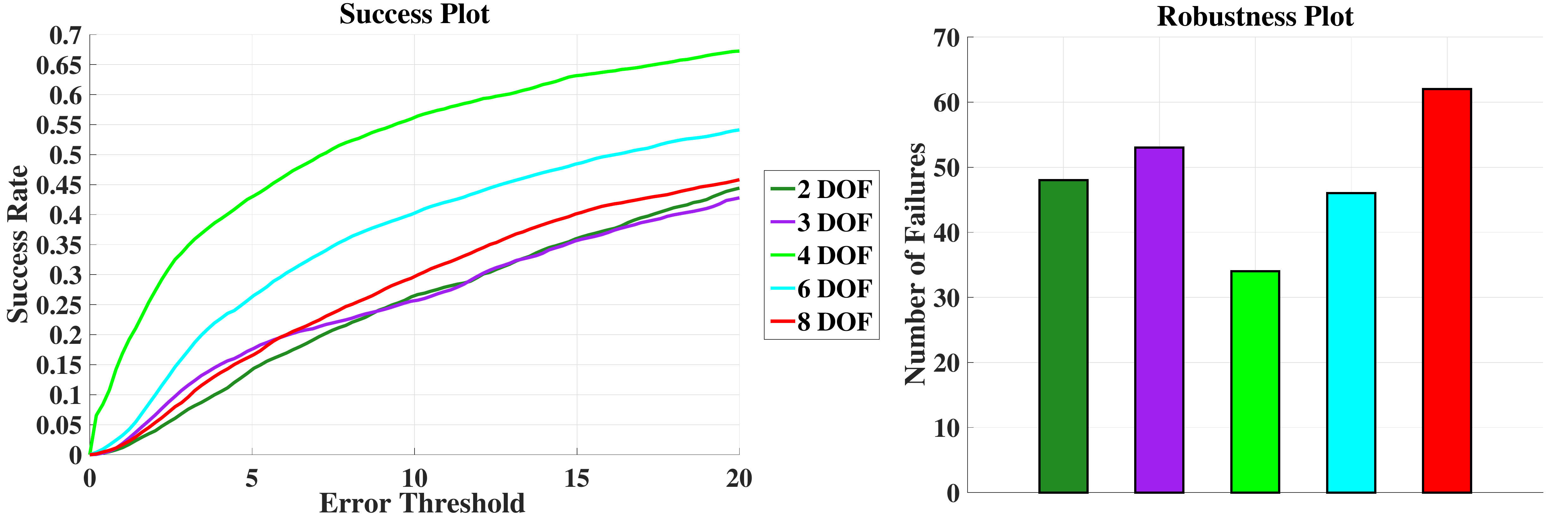}
\caption{Comparing different DoFs with RKLT on TFMT using alignment error, where 4 DoF seems to be the best compromise}
\label{fig:dofrklt}
\end{center}
\end{figure*}

\begin{figure*}[!htbp]
\begin{center}
\includegraphics[width=\textwidth]{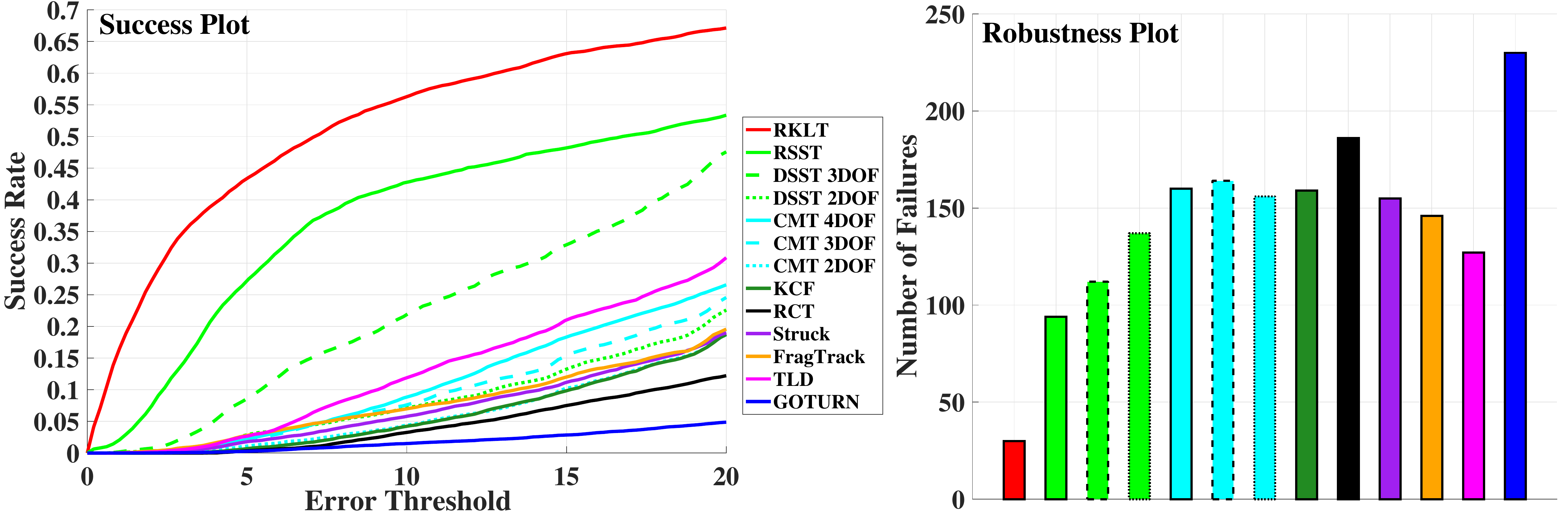}
\caption{Comparing different trackers on TFMT using alignment error}
\label{fig:tfmt_mcd}
\end{center}
\end{figure*}

\begin{figure}[!htbp]
\includegraphics[width=3.4in]{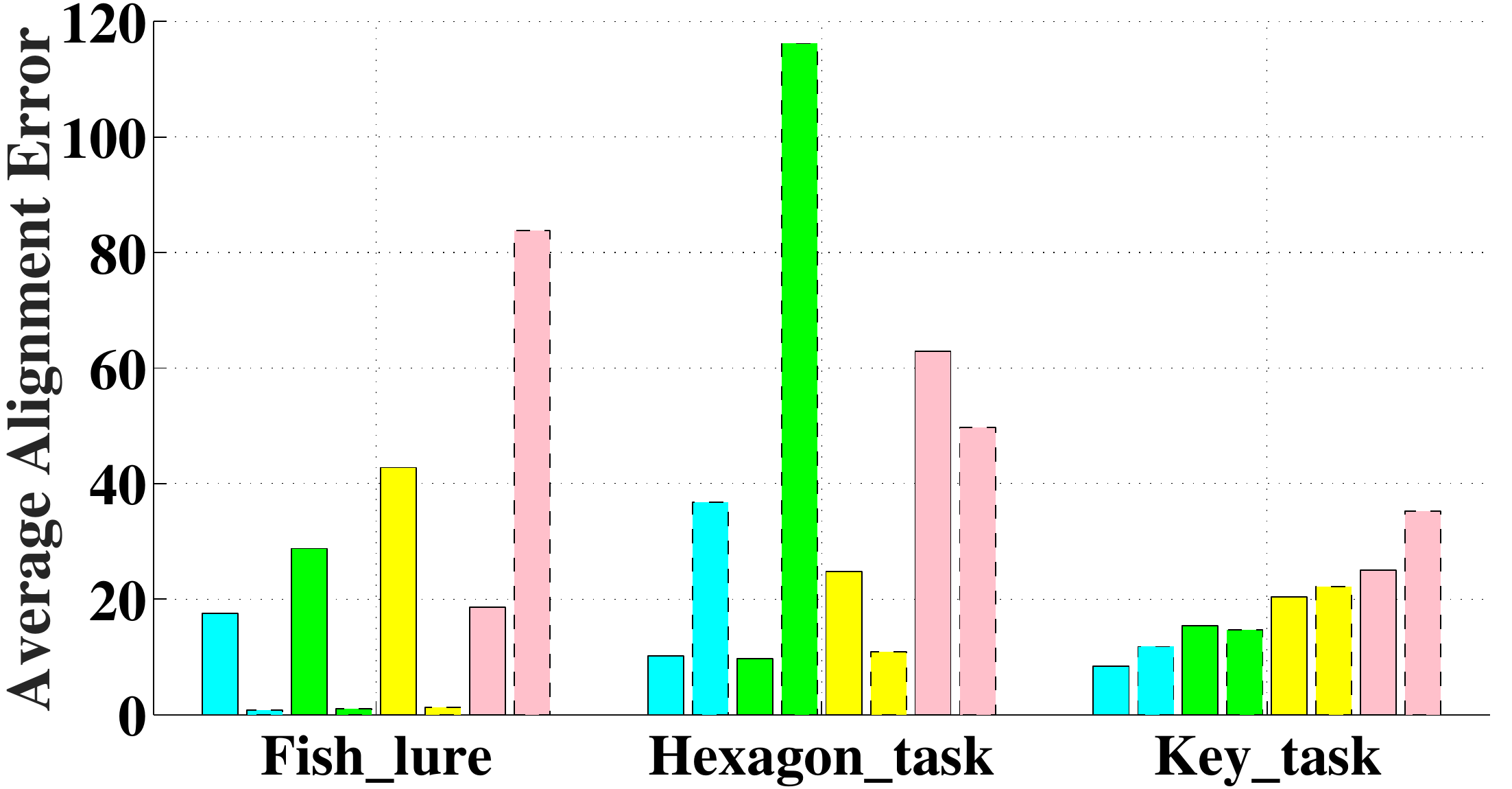}
\caption{Average Alignment Errors for RSST (solid edge) and RKLT (dotted edge) for different sequences in TFMT. Slow sequences from left and right camera are in \textcolor{cyan}{cyan} and \textcolor{green}{green} respectively while the corresponding fast sequences are in \textcolor{yellow}{yellow} and \textcolor{pink}{pink}.}
\label{fig:avg_mcd_err_tfmt}
\end{figure}
\begin{centering}
\begin{figure*}[!htbp]
\begin{subfigure}[b]{0.25\textwidth}
                \includegraphics[width=\linewidth]{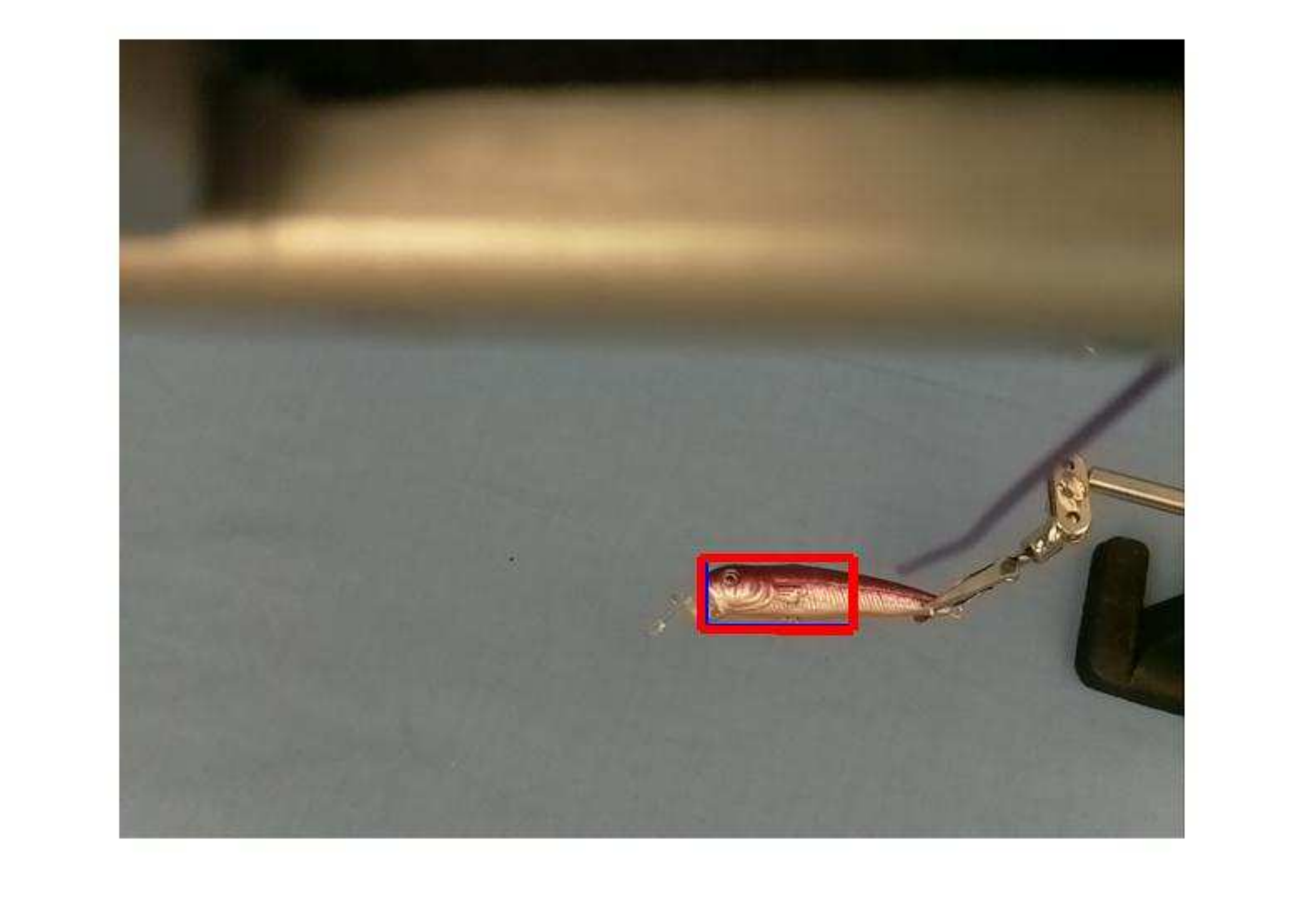}
        \end{subfigure}%
        \hspace{-1\baselineskip}
        \begin{subfigure}[b]{0.25\textwidth}
                \includegraphics[width=\linewidth]{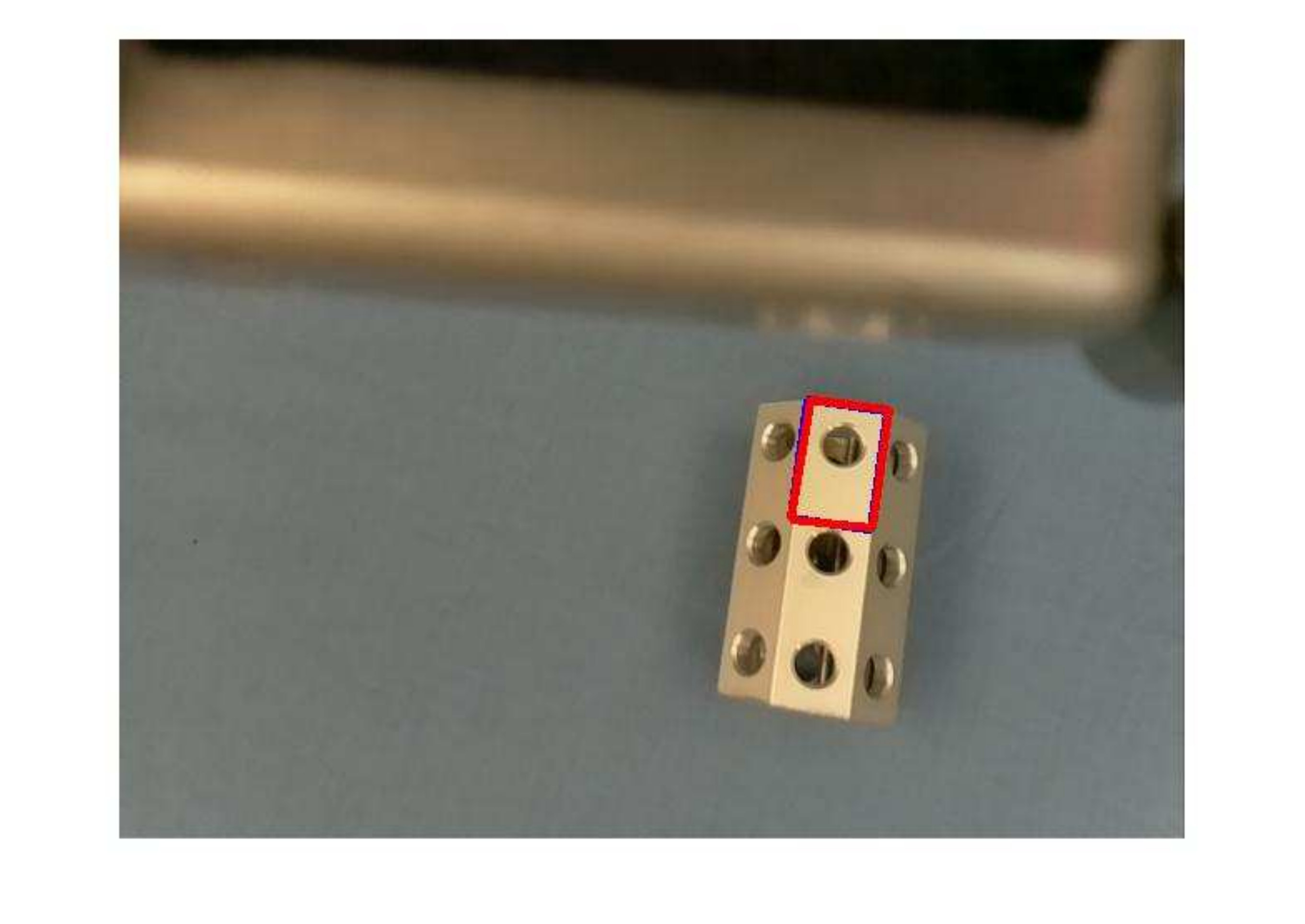}
        \end{subfigure}%
        \hspace{-1\baselineskip}
        \begin{subfigure}[b]{0.25\textwidth}
                \includegraphics[width=\linewidth]{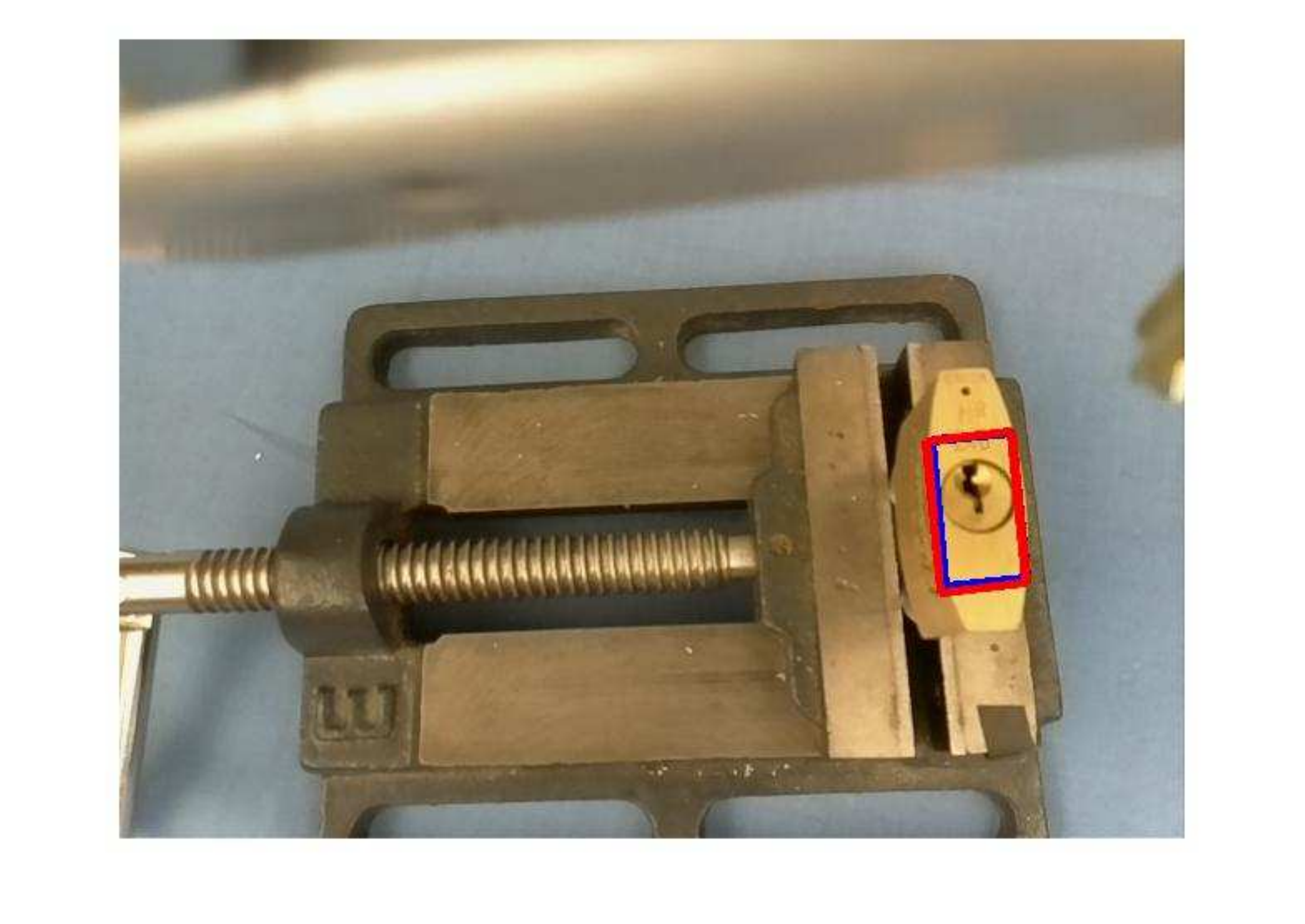}
        \end{subfigure}%
        \hspace{-1\baselineskip}
        \begin{subfigure}[b]{0.25\textwidth}
                \includegraphics[width=\linewidth]{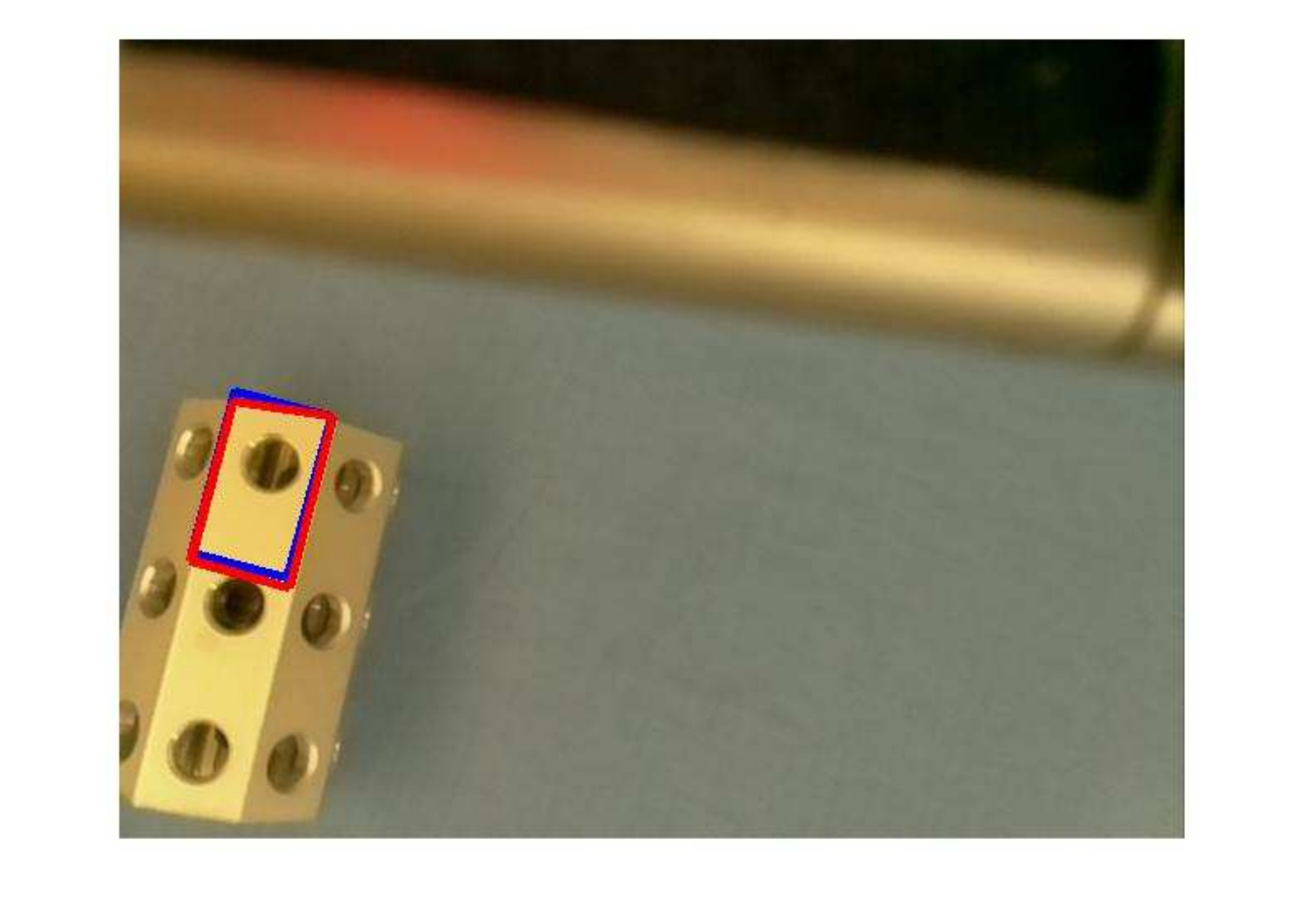}
        \end{subfigure}
        
        
        \vspace{-2\baselineskip}
        \begin{subfigure}[b]{0.25\textwidth}
          \includegraphics[width=\linewidth]{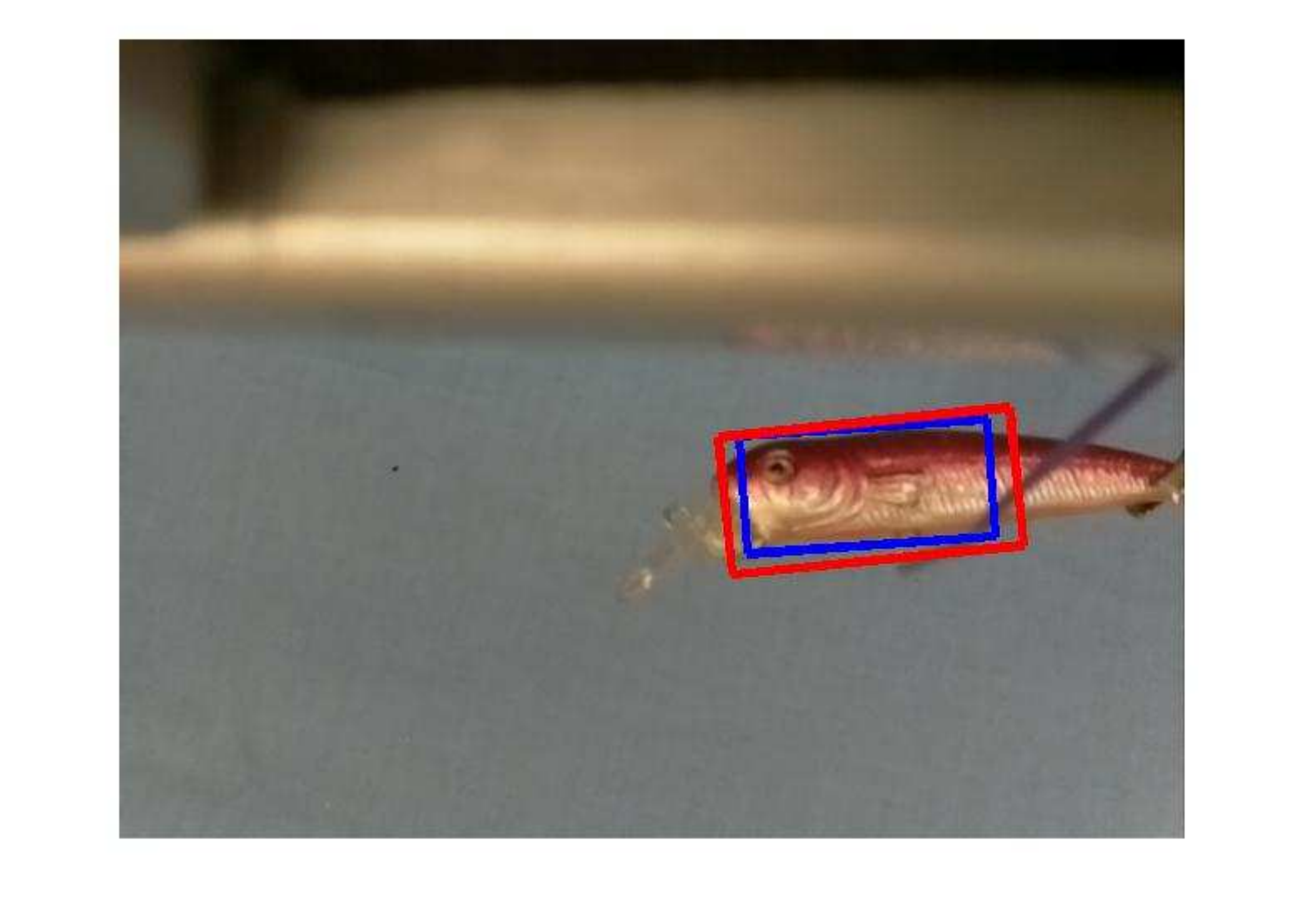}
        \subcaption{}
        \end{subfigure}%
        \hspace{-1\baselineskip}
        \begin{subfigure}[b]{0.25\textwidth}
                \includegraphics[width=\linewidth]{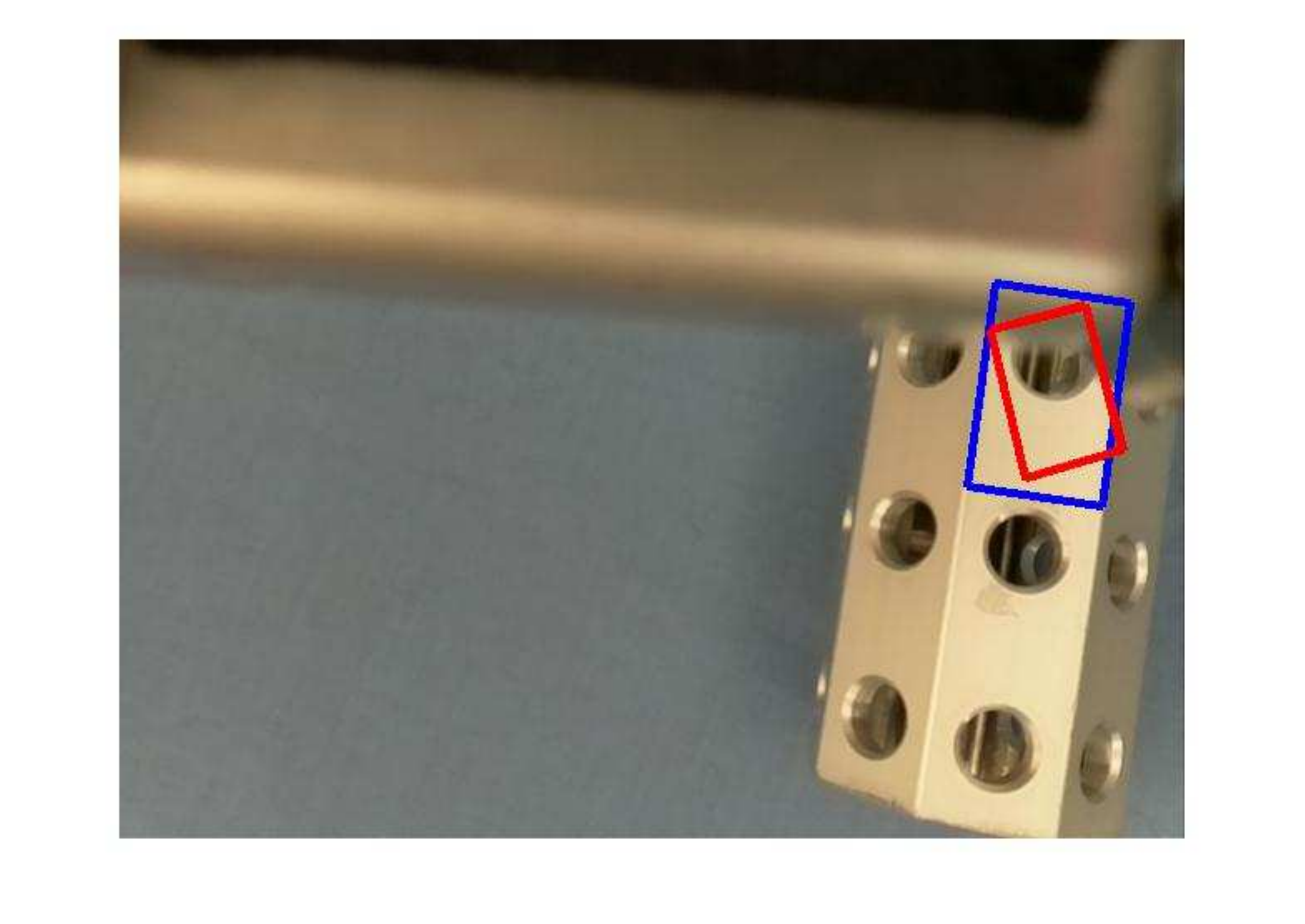}
        \subcaption{}
        \end{subfigure}%
        \hspace{-1\baselineskip}
        \begin{subfigure}[b]{0.25\textwidth}
                \includegraphics[width=\linewidth]{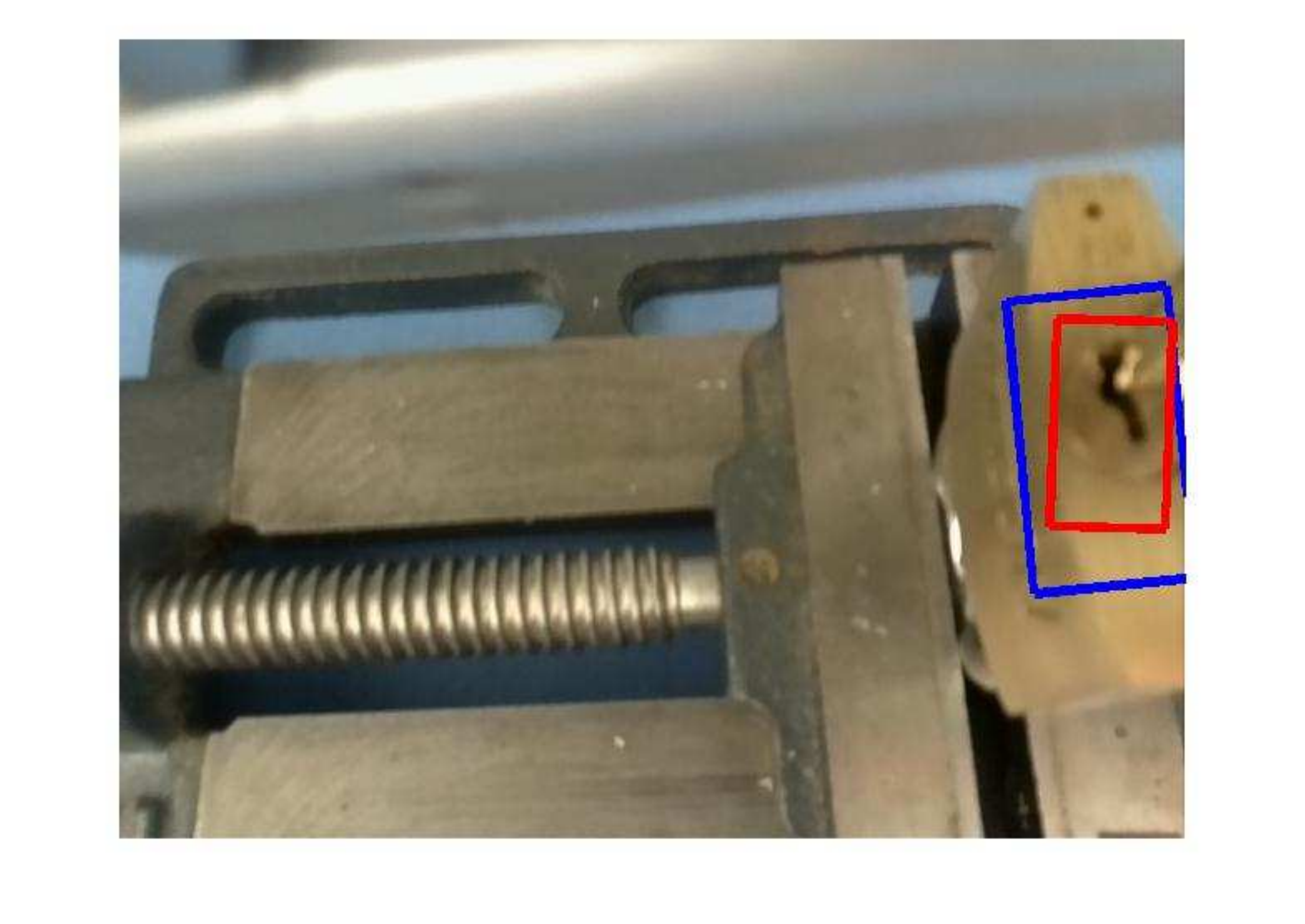}
        \subcaption{}
        \end{subfigure}%
        \hspace{-1\baselineskip}
        \begin{subfigure}[b]{0.25\textwidth}
                \includegraphics[width=\linewidth]{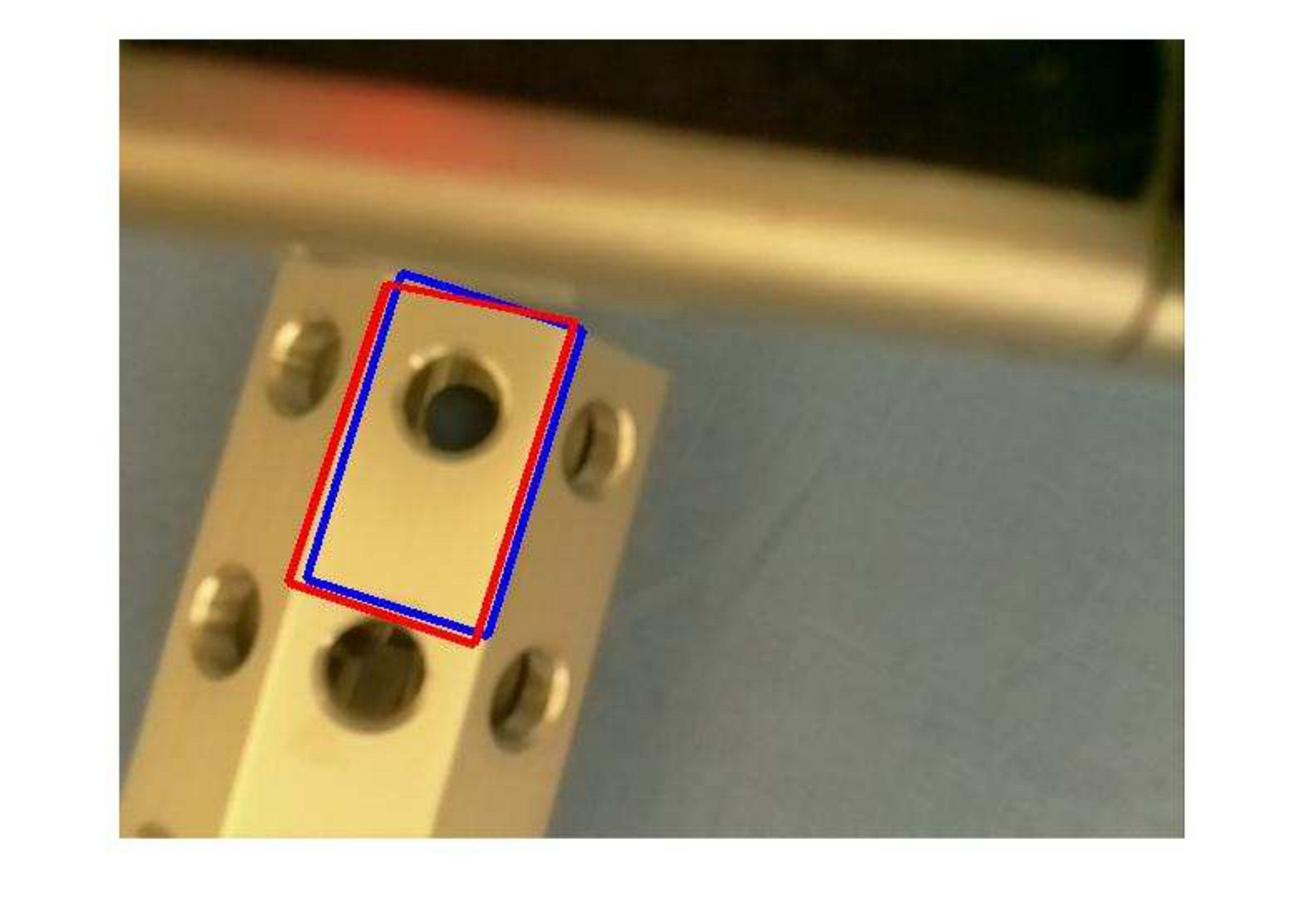}
        \subcaption{}
        \end{subfigure}
\caption{Tracking Results for RSST in blue and RKLT in red on TFMT. a: Fish Lure Left, b: Hexagon Task Left, c: Key Task Left, d: Hexagon Task Right.}
\label{fig:fine}
\end{figure*}
\end{centering}

\begin{figure*}[!htbp]
\begin{center}
\includegraphics[width=\textwidth]{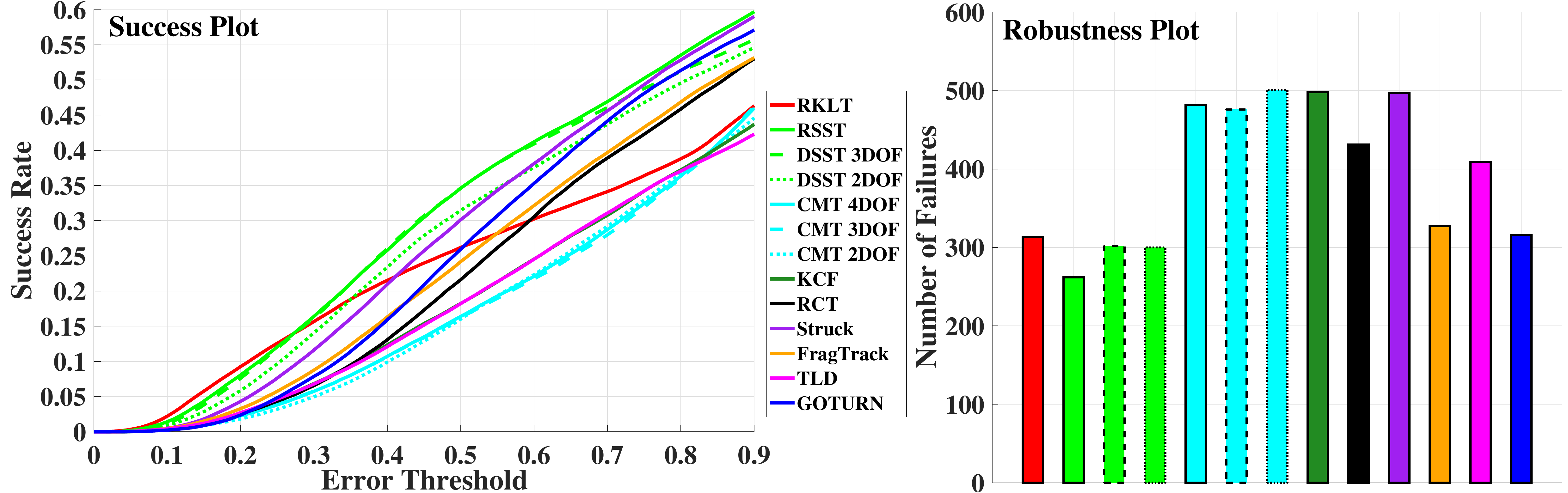}
\caption{Comparing different trackers on VOT 2016 using Jaccard error}
\label{fig:vot_jaccard}
\end{center}
\end{figure*}
Figure \ref{fig:tfmt_mcd} shows success and robustness results on TFMT where RKLT and RSST can again be seen to outperform all other trackers. However, since these are combined plots on all sequences, a more detailed comparison of RKLT and RSST is also performed on individual tasks. Figure \ref{fig:avg_mcd_err_tfmt} shows the average misalignment error on the four corners of the bounding box in pixels. The figure shows both left and right sequence alignment error for two speeds of performing the tasks. Images of the tracking results of both are shown in Figure \ref{fig:fine}. 
\par It is interesting to note that both trackers have an advantage in certain aspects. The RKLT tracker performs very well with high precision on fish lure which does not suffer as much from partial occlusions as the rest of the sequences.
This is expected since one of the strengths of registration based trackers is their accuracy.
On the other hand, hexagon task with normal speed and key task suffer a lot from partial occlusions and also have an object that is almost texture less. In this case, RSST is generally more robust than RKLT as shown by both the alignment error and the second column of the qualitative results.
That shows these two trackers complement each other and can be used for validating one another. 

\par Another finding that led us to use four DoF trackers is that low DoF trackers generally tend to be more stable as their search space is limited.
However, their overall tracking performance is only better when the actual object motion to be tracked does not significant exceed their capabilities.
This is shown in Figure \ref{fig:dofrklt}, where RKLT outperforms the 2, 3, 6 and 8 DoF versions where two DoF is simple 2D translation, three DoF includes isotropic scaling while four DoF adds on rotation. Six DoF uses affine transformation and eight degrees of freedom stands for using a homography.
The superiority of 4 DoF is apparent in both the robustness and success plots.
This can be explained because four DoF is the minimum to capture most motions that objects undergo. At the same time it's low enough to not have the gradient based methods get stuck in local minimas.


\subsection{General 2D Object Tracking}
Finally, an evaluation of these trackers in comparison to the state of the art on the VOT \cite{vot16} benchmark is presented in this section.
The reason for this is two fold.
The first reason is to show that RSST is still able to perform at par with the best tracker for the general 2D object tracking problem.
The second and more important reason is to demonstrate the shortcomings of VOT sequences for evaluating general 2D object tracking problem.
Figure \ref{fig:vot_jaccard} shows both the success and robustness plot. RSST is the best in terms of robustness while being slightly better than DSST in terms of success rate and outperforming the rest. On the other hand, RKLT does not perform well on this benchmark, except when using small error thresholds.

\par It is very interesting to see that one of the state of the art trackers, GOTURN \cite{held2016learning} that is based on deep regression networks, achieves very good results on VOT. However, it was the worst tracker on TMT. This is due to the fact that the tracker is trained offline with videos that are more similar to VOT sequences. These sequences are significantly different from the manipulation tasks scenarios.
It seems fair then to conclude that manipulation tasks and robotic scenarios in general offer different challenges than those present in VOT like benchmarks that are so popular in literature.

\section{Conclusions}
\label{sec:conc}
This paper introduced a new 4-DoF correlation based tracker (RSST) that can be used in the robotic fine manipulation context. A detailed analysis on manipulation tasks benchmark along with fine manipulation sequences and VOT benchmark was presented. This analysis showed that the strength of RSST lies in its ability to handle partial occlusions and objects going partially out of the field of view while still providing sufficiently precise results. This is the reason that it is the only tracker that is able to perform well on both the general 2D object tracking and manipulation tasks benchmarks.
A registration based tracker based on RANSAC and IC for four degrees of freedom was also presented and shown to perform competitively with RSST on the manipulation tasks.
It was also shown that four DoF tracking provides a good compromise between accuracy and robustness.

\par A new fully annotated dataset called tracking for fine manipulation tasks (TFMT) was presented with eye-in-hand camera configuration sequences. This complements the manipulation tasks (TMT) dataset with eye in hand sequences and different motion patterns.
Finally,  the differences between the general object tracking benchmarks and manipulation tasks benchmark were shown too.
This motivates the work for gathering sequences from real robot and human manipulation scenarios for testing such trackers.


\bibliographystyle{IEEEtran}
\bibliography{references}

\end{document}